\documentclass[11pt]{article}
\PassOptionsToPackage{table}{xcolor}
\usepackage{acl}
\usepackage{times}
\usepackage{latexsym}
\usepackage[T1]{fontenc}
\usepackage[utf8]{inputenc}
\usepackage{microtype}
\usepackage{inconsolata}
\usepackage{graphicx}
\usepackage{booktabs}
\usepackage{amsmath}
\usepackage{amssymb}
\usepackage{mathtools}
\usepackage{multirow}
\usepackage{array}
\usepackage{float}
\usepackage{xcolor}
\usepackage{colortbl}
\usepackage{url}
\usepackage{enumitem}
\usepackage{caption}
\usepackage{subcaption}
\usepackage{placeins}

\definecolor{stratbest}{RGB}{220, 237, 200}  %
\definecolor{overallbest}{RGB}{214, 234, 248} %
\definecolor{stratlow}{RGB}{255, 235, 238}    %
\definecolor{overalllow}{RGB}{255, 224, 210}  %
\definecolor{Human}{HTML}{2C3E50}
\definecolor{BeamSearch}{HTML}{E74C3C}
\definecolor{ContrastiveSearch}{HTML}{2ECC71}
\definecolor{TemperatureSampling}{HTML}{F39C12}
\definecolor{TopKSampling}{HTML}{9B59B6}
\definecolor{TopPSampling}{HTML}{E67E22}
\definecolor{repetitive}{RGB}{230, 225, 240}    %
\definecolor{erratic}{RGB}{245, 225, 215}       %
\definecolor{cellgreen}{RGB}{220, 237, 200}   %
\definecolor{cellgray}{RGB}{240, 240, 240}    %
\definecolor{cellyellow}{RGB}{255, 248, 220}  %
\definecolor{cellred}{RGB}{255, 230, 230}     %
\definecolor{darkblue}{rgb}{0, 0, 0.5}

\title{The Truncation Blind Spot: How Decoding Strategies\\ Systematically Exclude Human-Like Token Choices}

\author{
\begin{tabular}{@{}c@{\hspace{3.5em}}c@{}}
\textbf{Esteban Garcés Arias}$^{1,2*}$ &
\textbf{Nurzhan Sapargali}$^{1}$ \\[0.5em]
\textbf{Christian Heumann}$^{1}$ &
\textbf{Matthias A{\ss}enmacher}$^{1,2}$ \\[0.8em]
\multicolumn{2}{c}{\normalfont\mdseries
$^{1}$Department of Statistics, LMU Munich} \\
\multicolumn{2}{c}{\normalfont\mdseries
$^{2}$Munich Center for Machine Learning (MCML)} \\
\multicolumn{2}{c}{\normalfont\mdseries
$^{*}$Corresponding author:
\href{mailto:Esteban.GarcesArias@stat.uni-muenchen.de}
{\nolinkurl{Esteban.GarcesArias@stat.uni-muenchen.de}}}
\end{tabular}
}

\hypersetup{pdftitle={The Truncation Blind Spot: How Decoding Strategies Systematically Exclude Human-Like Token Choices}, pdfauthor={Esteban Garcés Arias; Nurzhan Sapargali; Christian Heumann; Matthias A{\ss}enmacher}}

\begin{document}

\maketitle

\begin{abstract}
Why does machine-generated text remain detectable? We investigate a mechanistic explanation at the decoding stage: standard strategies such as top-$k$ and nucleus sampling restrict generation to high-probability tokens, while human writers routinely choose contextually appropriate words from deeper in the model's probability distribution. Truncation makes a measurable share of these choices unreachable; we call this the \emph{truncation blind spot}. Across five open models and three domains, 8--18\% of human-selected tokens fall outside common truncation boundaries. Linguistic analysis further reveals disproportionate exclusion of content-word tokens. In a benchmark comprising 1.8 million machine generations, classifiers using only predictability and lexical diversity achieve mean AUC-ROC near 0.97, with substantial variation across decoding settings and strong transfer across generators. Probability-floor samplers substantially narrow the blind spot, demonstrating that the choice of truncation criterion matters for retaining human-used tokens. Together, these findings characterize a source of human--machine distributional mismatch and motivate decoding methods that preserve contextually appropriate low-probability choices while maintaining generation quality. Code and data are available at \url{https://github.com/EstebanGarces/human_vs_machine}.
\end{abstract}

\section{Introduction}
\label{sec:introduction}

The ability to distinguish human-written text from machine-generated content has become increasingly important as large language models (LLMs) are widely deployed in applications ranging from content creation to automated communication \citep{jawahar2020automatic, tang2024science}. While a substantial body of work has focused on detection methods, a more fundamental question remains insufficiently understood: \textit{why} is machine-generated text detectable in the first place? This paper investigates a mechanistic explanation grounded in the token selection of modern decoding strategies. Dominant approaches to open-ended generation, such as top-$k$ sampling \citep{fan2018hierarchical}, nucleus sampling \citep{holtzman2020how}, and contrastive search \citep{su2023contrastivesearchneedneural}, share one property: they restrict generation to high-probability regions of the model distribution through likelihood-based truncation.

Human text, in contrast, routinely contains words that are contextually precise yet rank low under a model's next-token distribution. We measure this directly: the median rank of a human-selected token is 1--2, while the mean is 50--155 (Section~\ref{sec:truncation_evidence}). Exclusion arises from the interaction between the predictive distribution and the decoding rule. When many continuations have positive probability, some contextually acceptable choices can have moderate or low rank even under an accurate model. Truncation by rank or cumulative mass then cuts into this spread-out set, and the writer's specific choice often falls below the cutoff. Accounts of speech production describe how communicative goals guide lexical selection \citep{levelt1989speaking}, while relevance theory emphasizes the interaction between communicative intentions and context \citep{sperber1986relevance}. We call the resulting tokens the \textbf{truncation blind spot}: tokens that human writers demonstrably use, that receive nonzero probability, and that standard decoding renders unreachable.

\begin{figure*}[t]
    \centering
    \includegraphics[width=0.92\textwidth]{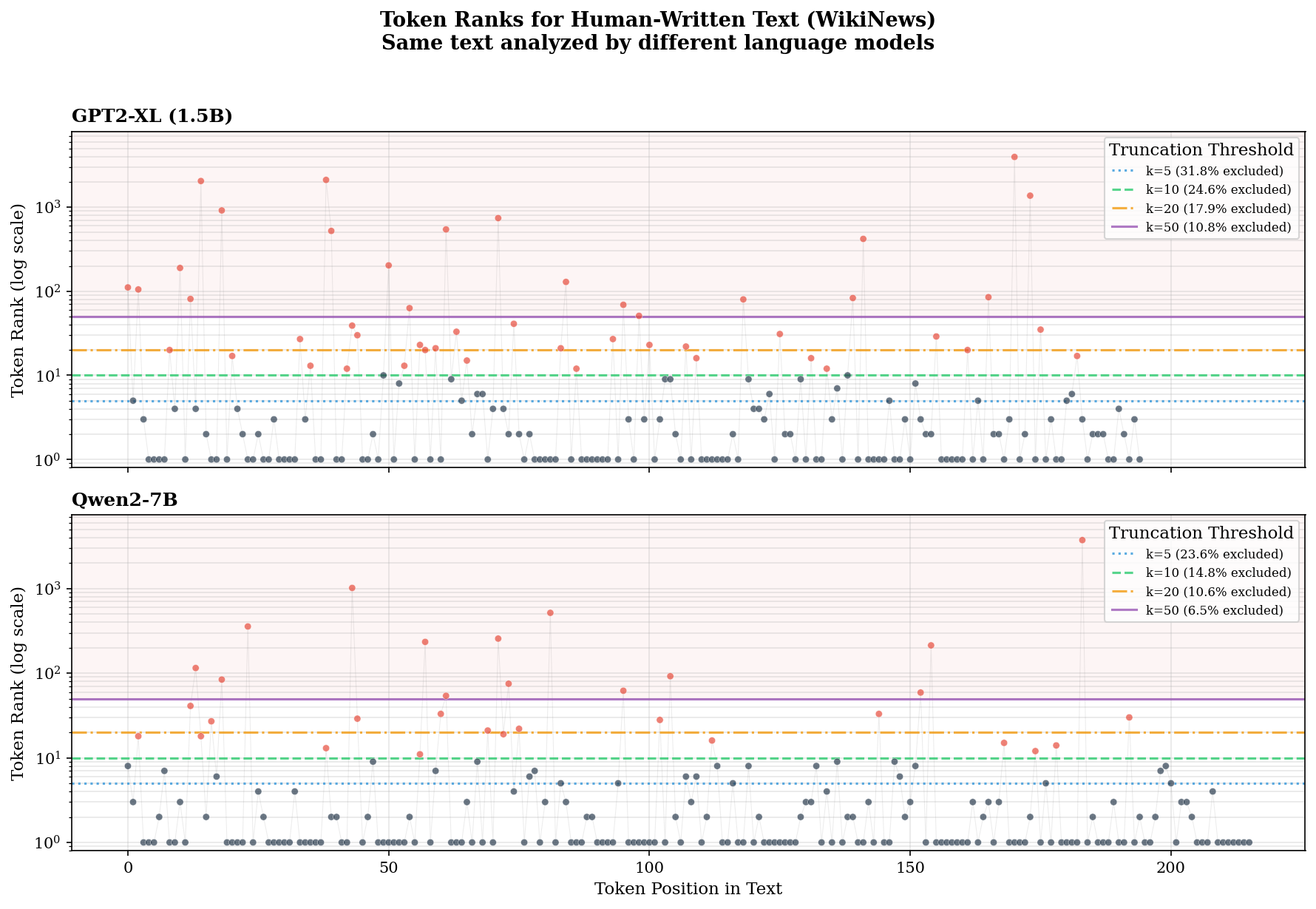}

    \scriptsize\textbf{From Wikinews:} \textit{``On 3 January 2026, the United States military invaded Venezuela, capturing and extrajudicially renditioning Venezuelan president Nicol\'as Maduro. [...] Maduro is to appear in a Manhattan federal court on 5 January 2026.''}
\caption{Token ranks (log scale) of a human-written WikiNews passage. Red marks tokens excluded under top-$k{=}10$, with per-model exclusion rates. Top: GPT2-XL; bottom: Qwen2-7B. The passage postdates all evaluated models' training cutoffs (Section~\ref{sec:truncation_evidence}).}
    \label{fig:decoding_behavior}
\end{figure*}

\noindent This mechanism offers one explanation for the reported convergence of different models toward similar outputs in open-ended settings \citep{jiang2025artificialhivemindopenendedhomogeneity}, since truncation constrains generation to similar regions of token space. We hypothesize that the same restriction induces systematic, detectable differences from human text, and investigate four questions:
\textbf{RQ1} (magnitude): to what extent does likelihood-based truncation exclude tokens that humans actually select? \textbf{RQ2} (detection signal): does this mismatch provide a reliable signal for distinguishing machine from human text? \textbf{RQ3} (mechanism primacy): is detectability governed by model characteristics or by the truncation mechanism? \textbf{RQ4} (quality trade-off): can configurations achieve low detectability while maintaining quality?

\noindent Since human token-selection distributions are not directly observable, we employ measurable proxies: \textit{predictability} (average log-likelihood under a reference model) and \textit{lexical diversity} ($n$-gram variety). We conduct a large-scale empirical study comprising 1,862,394 machine-generated texts from six open models, across five decoding strategies and 59 hyperparameter configurations, with supplementary proprietary and non-Transformer evaluations, alongside 5,261 human-written reference texts from three domains.

\noindent \paragraph{Contributions.} Building on studies of decoding and detectability \citep{ippolito2020automatic, dubois-etal-2025-sampling} and theoretical analyses of truncation \citep{hewitt2022truncation, finlayson2024closing}, we characterize the blind spot empirically and at scale: (i) a six-model detection grid with 59 decoding configurations and supplementary proprietary and non-Transformer evaluations; (ii) a part-of-speech and domain decomposition of the excluded tokens (Section~\ref{sec:pos_main}) and a within-family analysis of truncation-set overlap across scale (Appendix~\ref{sec:overlap}); (iii) comparisons across decoding settings, model scales, architectures, and proprietary systems, including cross-generator transfer with the generating model unseen at training time (Sections~\ref{sec:model_results} and~\ref{sec:cross_model}); and (iv) an evaluation of tail-recovering samplers against this benchmark: probability-floor criteria ($\epsilon$, $\eta$) yield lower exclusion than the nucleus baseline in all five open models at the least restrictive evaluated settings while typicality-based truncation does not; generation experiments test the exclusion--detectability link out of sample (Sections~\ref{sec:truncation_evidence}, \ref{sec:oos}). Together these characterize a decoding-related source of distributional mismatch, giving detection research a mechanism to investigate and decoding research a measurable cost to assess.

\section{Formal Setup: Truncation Sets and the Exclusion Rate}
\label{sec:preliminaries}

We define the objects our measurements operate on: the truncation set induced by a decoding strategy, the blind spot, and the exclusion rate that quantifies it. The quantity estimated in this formal setup is the exclusion rate of Equation~\ref{eq:exclusion_rate}, a sample statistic over observed human text that requires no model of the human production process.

\subsection{Likelihood-Based Truncation in Language Models}

Let $\mathcal{M}$ be an autoregressive language model with vocabulary $\mathcal{V}$. For each token $x_t \in \mathcal{V}$ at step $t$ given context $\mathbf{x}_{<t} = (x_1, \dots, x_{t-1})$, the model produces a categorical distribution $\mathbf{p}_t$ on the simplex $\Delta^{|\mathcal{V}|-1}$ with entries $P_\mathcal{M}(x_t \mid \mathbf{x}_{<t})$. For a truncation sampler, a \textit{support-selection rule} $\mathcal{S}: \Delta^{|\mathcal{V}|-1} \to 2^\mathcal{V} \setminus \{\emptyset\}$ maps $\mathbf{p}_t$ to a truncation set $\mathcal{T} \subseteq \mathcal{V}$, from which the next token is sampled (renormalized). Top-$k$ sampling retains the $k$ most probable tokens, $\mathcal{T}(\mathbf{x}_{<t}) = \{ x \mid \mathrm{rank}(x \mid \mathbf{x}_{<t}) \leq k \}$, where $\mathrm{rank}(x \mid \mathbf{x}_{<t}) = |\{ v \in \mathcal{V} : P_\mathcal{M}(v \mid \mathbf{x}_{<t}) \geq P_\mathcal{M}(x \mid \mathbf{x}_{<t}) \}|$, so that tied tokens share the rank of the last member of their tie group; nucleus (top-$p$) sampling retains the smallest top-ranked set with cumulative mass at least $p$; contrastive search uses a top-$k$ candidate pool but selects by a score that also penalizes degeneration. Beam search prunes competing prefixes by sequence score; it is included in our empirical comparisons but is not represented by this token-local sampling rule. We define a neural text generator as $\theta = (\mathcal{M}, \mathcal{S})$ with truncation set $\mathcal{T}_\theta(\mathbf{x}_{<t})$ and \emph{excluded region} $\mathcal{E}_\theta(\mathbf{x}_{<t}) = \mathcal{V} \setminus \mathcal{T}_\theta(\mathbf{x}_{<t})$, i.e., the tokens unreachable under $\theta$.

\subsection{The Blind Spot}
\label{sec:intent_driven}

One construct on the human side suffices: the \emph{human truncation set} $\mathcal{T}_{\mathcal{H}}(\mathbf{x}_{<t}) \subseteq \mathcal{V}$, the tokens a writer would accept as continuations at position $t$. It is a definitional device rather than an estimand: we never measure $\mathcal{T}_{\mathcal{H}}$ itself, only the event that a token a writer actually produced, by construction an element of $\mathcal{T}_{\mathcal{H}}$, falls outside the model's truncation set. Our analysis concerns the \emph{blind spot}, $\mathcal{T}_{\mathcal{H}} \cap \mathcal{E}_{\theta}$: tokens a writer selects that the model excludes. Appendix~\ref{sec:token_taxonomy} gives the full four-case taxonomy.

\paragraph{Prediction and reachability.} Even an accurate predictive distribution can assign positive probability to more acceptable continuations than a truncation rule retains. For example, if $m>k$ acceptable tokens have positive probability, a top-$k$ rule necessarily excludes some of them; a nucleus rule does so whenever its retained set omits an acceptable token. The blind spot is a property of the pair $\theta=(\mathcal{M},\mathcal{S})$: improving the model can change which tokens are excluded, while changing the decoding rule can change their reachability without changing the model.

\subsection{Quantifying the Blind Spot}

We quantify how often human-selected tokens fall into the blind spot. Since every observed human token lies in $\mathcal{T}_{\mathcal{H}}$ by construction, the blind-spot event coincides with exclusion by the model:
\begin{equation*}
    x^H_t \in \mathcal{T}_{\mathcal{H}}(\mathbf{x}^H_{<t}) \cap \mathcal{E}_{\theta}(\mathbf{x}^H_{<t})
    \iff
    x^H_t \notin \mathcal{T}_{\theta}(\mathbf{x}^H_{<t}).
\end{equation*}
Its magnitude is the rate at which human tokens fall outside the model's truncation set,
\begin{equation}
    \mathcal{R}_{\text{excl}}(\mathbf{x}^H; \theta) = \frac{1}{T} \sum_{t=1}^{T} \mathbf{1}\left[ x^H_t \notin \mathcal{T}_\theta(\mathbf{x}^H_{<t}) \right],
    \label{eq:exclusion_rate}
\end{equation}
where $\mathbf{1}[\cdot]$ is the indicator function: a direct sample statistic that counts how often a writer's token would have been unreachable under generator $\theta$, requiring no assumptions about the process that produced the text.

\subsection{Measurable Proxies}

Two text properties reflect the mismatch. \emph{Predictability}: machine text constrained to high-probability tokens should show higher average log-likelihood under a reference model than human text. \emph{Lexical diversity}: by concentrating mass on frequent tokens, truncation should reduce $n$-gram variety relative to human text, which draws on a broader effective vocabulary.

\section{Related Work}
\label{sec:related_work}
Detection research spans supervised classifiers \citep{jawahar2020automatic, tang2024science} and zero-shot statistical methods \citep{mitchell2023detectgpt}. \citet{gehrmann2019gltr} visualized that machine tokens concentrate in high-probability regions; we quantify the converse, how often human tokens fall outside truncation sets. Closest to our work, \citet{ippolito2020automatic} showed that decoding choices affect detectability, and \citet{dubois-etal-2025-sampling} report that sampling parameters move detector accuracy. We add the token-level mechanism behind these observations: a measured exclusion rate with a known linguistic composition, evidence relating this mechanism to detectability across models, and an evaluation of proposed remedies against it.

On the decoding side, truncation samplers trade quality against diversity \citep{fan2018hierarchical, holtzman2020how, su2023contrastivesearchneedneural, garces-arias-etal-2025-decoding}. \citet{hewitt2022truncation} cast truncation as desmoothing and propose $\eta$- and $\epsilon$-sampling; \citet{meister2023locally} select tokens by typicality. We take the complementary, diagnostic view, asking what truncation excludes relative to human text, and we evaluate these samplers against our exclusion benchmark (Sections~\ref{sec:truncation_evidence} and~\ref{sec:oos}). Adaptive Contrastive Search adjusts its degeneration penalty using uncertainty \citep{garces-arias-etal-2024-adaptive}, and GUARD combines local and global entropy signals \citep{ding-etal-2025-guard}. Min-$k$ selects truncation boundaries from relative logit changes \citep{ding-etal-2026-min}. These methods offer further candidates for evaluating human-token retention. Psycholinguistic and information-theoretic accounts of language \citep{levelt1989speaking, jaeger2010redundancy} provide a complementary perspective on human token choice (extended discussion in Appendix~\ref{app:extended_rw}).
\section{Experimental Setup}
\label{sec:experimental_setup}

\subsection{Language Models}

We evaluate eight language models spanning sizes, architectures, and training procedures: the open-source Llama-3.1 \citep{touvron2023llama}, Mistral-7B \citep{jiang2023mistral}, Qwen 2 \citep{bai2023qwen}, Falcon 2 \citep{almazrouei2023falcon}, Deepseek \citep{liu2024deepseek}, and GPT2-XL \citep{radford2019language}, and the proprietary GPT-3.5-turbo and Claude-3-Haiku. We additionally include the non-Transformer architectures Mamba \citep{gu2024mambalineartimesequencemodeling} and RWKV \citep{peng-etal-2023-rwkv} to test whether the phenomenon is Transformer-specific. Proprietary systems are queried under provider API defaults, the conditions under which they are deployed; the exclusion analyses use the checkpoints listed in Appendix~\ref{app:checkpoints}.

\subsection{Decoding Strategies and Configurations}
\label{sec:decoding_setup}

Following \citet{garces-arias-etal-2025-decoding}, we evaluate five decoding strategies: beam search (widths $\{3, 5, 10, 15, 20, 50\}$), contrastive search \citep{su2023contrastivesearchneedneural} ($k \in \{1, 3, 5, 10, 15, 20, 50\}$, $\alpha \in \{0.2, \dots, 1.0\}$), top-$k$ sampling \citep{fan2018hierarchical} ($k \in \{1, 3, 5, 10, 15, 20, 50\}$), top-$p$ sampling \citep{holtzman2020how} ($p \in \{0.6, 0.7, 0.8, 0.9, 0.95\}$), and temperature sampling ($T \in \{0.1, 0.3, 0.5, 0.7, 0.9, 1.0\}$), giving 59 configurations per model and 1,862,394 generated texts across the six open models. The four non-temperature strategies contribute 53 configurations. Temperature rescales probabilities without itself removing support; inherited sampling defaults can impose additional truncation. In Section~\ref{sec:truncation_evidence} we additionally evaluate three tail-recovering truncation rules as \emph{exclusion criteria} on human text: $\epsilon$-sampling ($\epsilon \in \{3, 9, 20\}\times10^{-4}$) and $\eta$-sampling ($\epsilon \in \{0.9, 2, 4\}\times10^{-3}$; both rules are parameterized by $\epsilon$ in \citealp{hewitt2022truncation}), and locally typical sampling ($\tau \in \{0.2, 0.9, 0.95\}$) \citep{meister2023locally}; Section~\ref{sec:oos} additionally generates with a subset of these rules for two models. Grid generation follows the generation library's default settings, under which the truncation applied in total can be stricter than the nominal parameter alone; the harness of Section~\ref{sec:oos} instead applies each rule in isolation.

\subsection{Datasets}

We employ three datasets representing distinct writing styles, totaling 5,261 human-written texts: BookCorpus \citep{zhu2015aligning} (1,947 samples of creative fiction), WikiText \citep{merity2016pointer} (1,314 Wikipedia articles), and WikiNews\footnote{\url{http://www.wikinews.org}} (2,000 news reports).

\subsection{Metrics}
\label{sec:metrics}

\paragraph{Predictability.} Given a text $\mathbf{x} = (x_1, \ldots, x_T)$, prompt $\mathbf{c}$, and reference model $\mathcal{M}_{\text{ref}}$:
\begin{equation}
    \text{Pred}(\mathbf{x} \mid \mathbf{c}) = \frac{1}{T} \sum_{t=1}^{T} \log P_{\mathcal{M}_{\text{ref}}}(x_t \mid [\mathbf{c} : \mathbf{x}_{<t}]).
    \label{eq:predictability}
\end{equation}
Higher (less negative) values indicate higher likelihood; predictability is the negative of mean per-token log-perplexity, and we use the term for accessibility. We use OPT-2.7B \citep{zhang2022optopenpretrainedtransformer} as $\mathcal{M}_{\text{ref}}$, following the evaluation setup of \citet{garces-arias-etal-2025-statistical}; results are robust to the choice of judge model (pairwise $\rho \geq 0.88$; Appendix~\ref{sec:robustness}). 

\paragraph{Lexical Diversity.} We measure $n$-gram diversity as
\begin{equation}
    \text{Div}(\mathbf{x}) = 100 \cdot \prod_{n=2}^{4} \frac{|\text{unique } n\text{-grams}(\mathbf{x})|}{|\text{total } n\text{-grams}(\mathbf{x})|}.
\end{equation}
Low diversity indicates repetition; high diversity can reflect genuine richness or incoherent randomness. The two features enter all analyses separately: low entropy at generation time need not imply low lexical diversity, and the two properties capture complementary aspects of the mismatch.

\subsection{Classification Setup}
\label{sec:classification}

We train binary classifiers, human-written versus machine-generated text from a given (model, configuration) cell, on the two features alone: Random Forest \citep{breiman2001random} (RF, 500 trees), Logistic Regression (LR), and Naive Bayes (NB), with 80-20 class-stratified splits of individual texts for the per-configuration grid. Human and machine continuations from the same prompt can occur in different partitions. Simple classifiers are intentional: high performance with two features indicates systematically different statistics. AUC-ROC is invariant to the class prior, and downsampling machine examples leaves it essentially unchanged (Appendix~\ref{sec:class_balance}).

\section{Results}
\label{sec:results}

\subsection{The Magnitude of the Blind Spot (RQ1)}
\label{sec:truncation_evidence}

\begin{table}[t]
\centering
\small
\setlength{\tabcolsep}{4pt}
\begin{tabular}{lcccc}
\toprule
\textbf{Setting} & \textbf{Book} & \textbf{News} & \textbf{Wiki} & \textbf{Avg.\ (95\% CI)} \\
\midrule
$k=5$  & 25.0 & 22.5 & 27.4 & 25.0 [23.1, 27.4] \\
$k=10$ & 18.1 & 15.7 & 20.4 & 18.1 [16.3, 20.2] \\
$k=20$ & 12.9 & 10.8 & 15.2 & 13.0 [11.4, 14.8] \\
$k=50$ & 8.0  & 6.5  & 9.9  & 8.1 [6.9, 9.5] \\
\midrule
$p=0.6$  & 28.1 & 28.5 & 29.6 & 28.7 [27.7, 29.8] \\
$p=0.9$  & 7.4  & 8.7  & 7.9  & 8.0 [7.4, 8.8] \\
$p=0.95$ & 3.5  & 4.5  & 3.7  & 3.9 [3.4, 4.6] \\
\bottomrule
\end{tabular}
\caption{Human token exclusion rates $\mathcal{R}_{\text{excl}}$ (\%) at representative top-$k$ and top-$p$ settings, averaged over five open models (GPT2-XL, Qwen2-7B, Mistral-7B-v0.3, Llama-3-8B, Falcon2-11B) and computed within each model's own tokenizer. Full sweep in Table~\ref{tab:exclusion_rates} (Appendix~\ref{sec:exclusion_details}).}
\label{tab:exclusion_main}
\end{table}

Figure~\ref{fig:decoding_behavior} illustrates the blind spot and Table~\ref{tab:exclusion_main} quantifies it: at $k{=}10$, roughly 18\% of human-selected tokens are unreachable; at $k{=}50$, roughly 8\%; nucleus sampling at $p{=}0.9$ excludes about 8\%. Exclusions concentrate on content words in our separate GPT-2 analysis (Section~\ref{sec:pos_main}); token-level rates alone do not determine passage-level reachability because exclusions can be dependent. The full sweep and rank distributions are in Appendix~\ref{sec:exclusion_details}. This is the signature of the mechanism in Section~\ref{sec:intent_driven}: at most positions the writer's token is where the model expects it, but at positions of lexical freedom the model's mass is spread over many plausible candidates and fixed cutoffs remove the chosen one. WikiText shows the heaviest tail, consistent with specialized vocabulary. The Figure~\ref{fig:decoding_behavior} passage postdates every model's training cutoff, and its exclusion rates (24.6\%/14.8\% at $k{=}10$) are comparable to the WikiNews average (15.7\%), showing that exclusion also occurs in text written after pretraining.

\paragraph{Tail-recovering truncation criteria.}
Our account makes a testable prediction: truncation criteria that impose no fixed rank or mass budget should retain more of the human-used tail at matched permissiveness. We evaluate the exclusion side of this prediction for $\eta$- and $\epsilon$-sampling \citep{hewitt2022truncation} and locally typical sampling \citep{meister2023locally} on the same human corpora, for the five open models of Table~\ref{tab:exclusion_main} (Table~\ref{tab:tail_samplers}). Because average truncation-set sizes are tokenizer-dependent, we compare within each model.

Probability-floor criteria confirm the prediction, and more strongly than matched permissiveness requires. In every model, both $\epsilon$- and $\eta$-sampling admit a setting that strictly dominates nucleus sampling at $p{=}0.9$, with lower exclusion at a smaller average truncation set and relative reductions of 23--38\% across all ten (model, rule) pairings (e.g., Falcon2-11B: $\epsilon$-sampling at 6.2\% exclusion and 39 tokens vs.\ nucleus at 10.1\% and 53). Aggregated over models and domains, exclusion falls from 8.8\% (nucleus, $p{=}0.9$) to 5.3\% ($\eta$) and 4.2\% ($\epsilon$) at comparable retained mass (Table~\ref{tab:tail_samplers}). Throughout the table, exclusion closely tracks the truncated probability mass ($\mathcal{R}_{\text{excl}} \approx 100 - \text{retained mass}$): human choices occupy the model's mass in proportion to it, an aggregate correspondence that does not, by itself, establish model calibration or rule out model error (Section~\ref{sec:intent_driven}).

Locally typical sampling, by contrast, tracks nucleus sampling within $0.1$ points in every model at matched mass ($\tau{=}0.9$: $8.8\%$ for both; $\tau{=}0.95$: $4.7\%$ for both) and at low $\tau$ \emph{amplifies} the blind spot ($\tau{=}0.2$: $54.9\%$), since the human-used tail consists precisely of the high-surprisal tokens that typicality treats as atypical. The operative distinction is therefore not likelihood-based versus information-based selection, but whether the criterion caps retained mass or rank (nucleus, top-$k$, typicality) or sets a per-token probability floor ($\epsilon$, $\eta$) that widens the pool where the distribution is flat, exactly the positions of greatest lexical freedom identified in Section~\ref{sec:intent_driven}. Even the best criterion leaves $4$--$5\%$ of human tokens unreachable: probability-floor truncation narrows the blind spot substantially but does not close it. Removing truncation restores support but does not, by itself, ensure generation quality. Section~\ref{sec:oos} tests this prediction directly by generating with these rules.

\begin{table}[t]
\centering
\small
\begin{tabular}{llccc}
\toprule
\textbf{Rule} & \textbf{Param} & $\mathcal{R}_{\text{excl}}$ (\%) & $\overline{|\mathcal{T}|}$ & Retained mass \\
\midrule
top-$k$   & $k=10$              & 18.8 & 10  & 0.82 \\
top-$k$   & $k=50$              & 8.1  & 50  & 0.92 \\
top-$p$   & $p=0.6$             & 29.6 & 7   & 0.72 \\
top-$p$   & $p=0.9$             & 8.8  & 86  & 0.92 \\
top-$p$   & $p=0.95$            & 4.7  & 211 & 0.96 \\
\midrule
$\epsilon$ & $3\times10^{-4}$   & 4.2  & 91  & 0.96 \\
$\epsilon$ & $9\times10^{-4}$   & 6.7  & 44  & 0.94 \\
$\eta$     & $9\times10^{-4}$   & 5.3  & 89  & 0.95 \\
$\eta$     & $2\times10^{-3}$   & 6.8  & 66  & 0.94 \\
\midrule
typical   & $\tau=0.2$          & 54.9 & 8   & 0.46 \\
typical   & $\tau=0.9$          & 8.8  & 91  & 0.92 \\
typical   & $\tau=0.95$         & 4.7  & 212 & 0.96 \\
\bottomrule
\end{tabular}
\caption{Human-token exclusion under tail-recovering rules versus rank and mass baselines, averaged over the five open models of Table~\ref{tab:exclusion_main} (checkpoints in Appendix~\ref{app:checkpoints}); computed within each model's tokenizer, conditioned on the prompt, ties per Section~\ref{sec:preliminaries}. The $k{=}10$ baseline is 18.8\% in this analysis and 18.1\% in Table~\ref{tab:exclusion_main}. Set sizes are tokenizer-dependent; dominance comparisons are made within each model (see text).}
\label{tab:tail_samplers}
\end{table}

\subsection{From Exclusion to Detectability: An Out-of-Sample Test}
\label{sec:oos}

The preceding analysis measures what these rules exclude; we now test whether the exclusion ordering predicts detectability for rules the detection grid has never seen. For GPT2-XL and Qwen2-7B we generate continuations of all 5,261 prompts under the four floor settings of Table~\ref{tab:tail_samplers}, under locally typical sampling ($\tau{=}0.9$), and under nucleus sampling ($p{=}0.9$) as baseline, each rule applied in isolation, scoring each configuration with the classifiers of Section~\ref{sec:classification}. We evaluate two predictions: (P1) AUC follows the exclusion ordering of Table~\ref{tab:tail_samplers}, placing lower-exclusion criteria below nucleus sampling; (P2) typical sampling at matched mass is indistinguishable from nucleus sampling.

\begin{figure}[t]
  \centering
  \includegraphics[width=0.98\columnwidth]{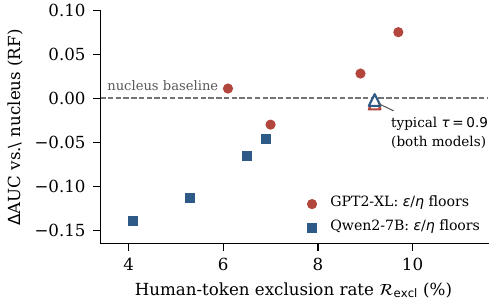}
  \caption{Exclusion and detectability for unseen sampling rules. Change in RF AUC-ROC relative to each model's nucleus baseline (dashed line at $\Delta{=}0$) for each out-of-sample configuration (Table~\ref{tab:oos_detectability}), against its human-token exclusion rate. Filled: probability-floor settings; open: typical sampling, close to the baseline.}
  \label{fig:auc_vs_exclusion}
\end{figure}

Figure~\ref{fig:auc_vs_exclusion} reports each configuration's $\Delta$AUC against its model's nucleus baseline (Table~\ref{tab:oos_detectability}, Appendix~\ref{app:oos_table}). Qwen2-7B supports P1: exclusion and AUC have perfect rank agreement across the evaluated configurations (Spearman $\rho=1.00$), and the strongest floor setting reduces AUC by 0.14 (RF) to 0.16 (LR). GPT2-XL provides mixed support: rank agreement varies by classifier ($\rho=0.37$--$0.83$), and some lower-exclusion settings increase AUC. For example, $\epsilon=3\times10^{-4}$ lowers exclusion from 9.2\% to 6.1\% but increases RF AUC by 0.011. P1 is therefore not consistently confirmed for GPT2-XL. For P2, typical sampling differs from nucleus sampling by at most 0.006 AUC in both models and the reported RF/LR comparisons. These small observed differences are consistent with P2.

These results provide model-dependent support for a relationship between exclusion and detectability. Probability-floor criteria substantially reduce detectability for Qwen2-7B, while GPT2-XL shows a mixed response. Its scope is two models, two detection features, and no quality evaluation of floor-generated text; \citet{hewitt2022truncation} report separate human evaluations favoring $\eta$-sampling at comparable settings.

\subsection{Linguistic Composition of the Blind Spot}
\label{sec:pos_main}

\begin{figure*}[t]
  \centering
  \includegraphics[width=0.95\textwidth]{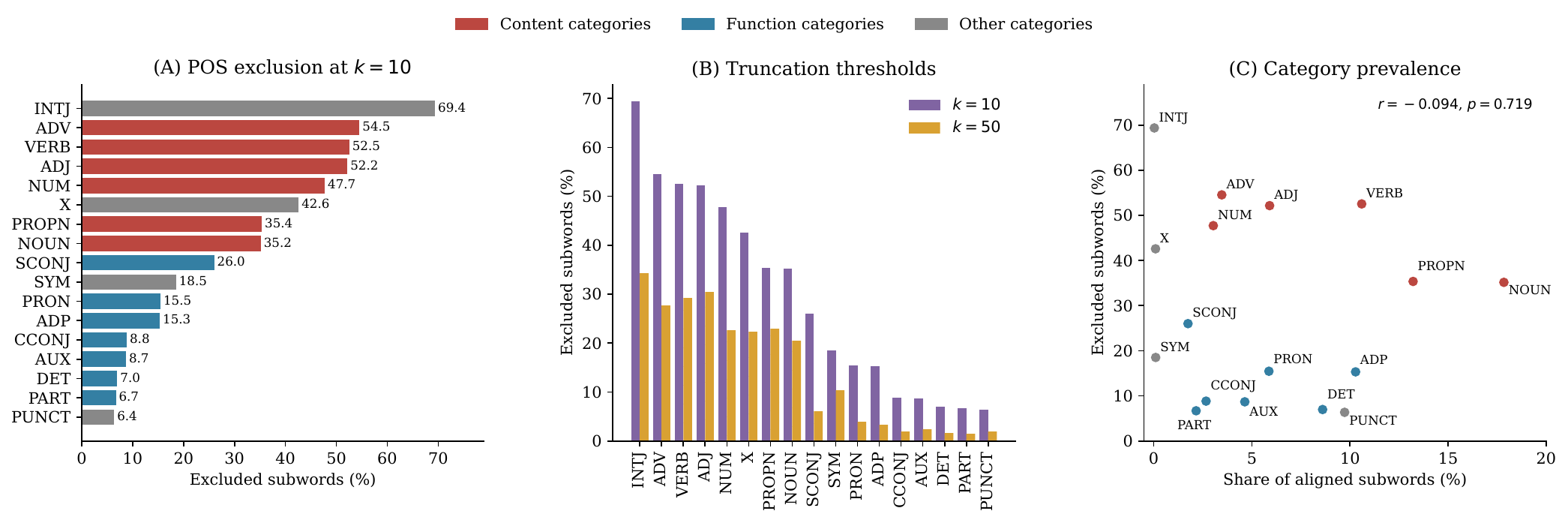}
  \caption{POS analysis of continuation subwords from 600 prompt/reference pairs, pooled across three GPT-2 checkpoints. \textbf{(A)} Exclusion at $k=10$: content categories (red), function categories (blue), and other tags (gray). \textbf{(B)} Exclusion at $k=10$ (purple) and $k=50$ (gold, hatched). \textbf{(C)} POS-category prevalence versus exclusion ($r=-0.09$). Rates count aligned subwords, not whole words.}
  \label{fig:pos_analysis}
\end{figure*}

We examine which tokens are excluded in a within-family analysis of GPT-2 small, medium, and large. A fixed sample contains 200 unique prompt/reference pairs per domain (600 total) from the benchmark's available reference checkpoints. We tag each prompt and continuation with spaCy \citep{honnibal2020spacy} and align model subwords to Universal POS tags \citep{nivre2016universal} using character offsets. Exclusion is measured on up to 256 continuation subwords, conditioned on the prompt. The shared GPT-2 tokenizer permits comparisons across model sizes without changing subword units. Appendix~\ref{sec:pos_analysis} details sampling, alignment, and aggregation.

At $k=10$, pooled exclusion is 42.4\% for subwords aligned to content words and 12.0\% for those aligned to function words: a ratio of 3.53 (95\% stratified prompt/reference-pair bootstrap CI: 3.43--3.64). The content--function ordering holds in all nine model/domain combinations (ratios 3.27--3.84). Category-specific rates range from 35.2--54.5\% for content categories and 6.7--26.0\% for function categories (Figure~\ref{fig:pos_analysis}A). At $k=50$, both pooled rates decrease, to 24.4\% and 2.7\%, respectively.

\subsection{Detection Signals Across Models and Decoding Settings (RQ2, RQ3)}
\label{sec:model_results}
\label{sec:predictability_results}

Using only predictability and diversity, all three classifiers achieve mean AUC-ROC of 0.968--0.974 and mean F1 of 0.949--0.953 across the 1{,}062 (model, domain, decoding configuration) cells of the grid (Table~\ref{tab:overall_performance}, Appendix~\ref{sec:model_details}). Strong performance across these classifiers shows that the two features carry substantial detection signal.

Detectability varies with both the evaluated models and their decoding settings (Appendix~\ref{sec:model_details}). Llama-3.1 has the highest mean RF AUC (0.993), while Falcon 2 has the lowest (0.945); these rankings do not follow model age or size alone. However, inherited generation defaults confound a comparison of model identity at a nominally fixed decoding setting (Section~\ref{sec:decoding_setup}). Within-family scale coefficients are non-significant for three of four families, but GPT-2 shows a significant negative effect ($\beta=-0.052$, $p=0.015$), as does the pooled estimate ($\beta=-0.021$, $p=0.017$). Non-Transformer systems have higher detectability in the architecture analysis ($\Delta$AUC $=+0.180$). High detectability therefore persists across the evaluated architectures, with measurable variation by model characteristics.

A within-family analysis of the truncation sets points the same way: models of one family share substantially overlapping sets (mean Jaccard at $k{=}10$: 0.589 for GPT-2, 0.531 for Qwen2), and overlap decreases only gradually with the size gap. These overlaps quantify shared candidates across model sizes (Appendix~\ref{sec:overlap}). Truncation parameters, by contrast, show a clear relationship with detectability: beam search is most detectable (AUC-ROC 0.997), followed by contrastive search (0.981), top-$p$ (0.956), and top-$k$ (0.948), a 4.9-point spread among these strategy averages (Table~\ref{tab:strategy_performance}, Appendix~\ref{app:strategy_table}). The strategy averages summarize different parameter grids and include all effects of each decoding configuration. For top-$k$, increasing $k$ from 1 to 15 reduces AUC-ROC from 0.988 to 0.924; for nucleus sampling, increasing $p$ from 0.6 to 0.95 reduces AUC-ROC from 0.978 to 0.935. Even the least detectable configurations do not fully match human statistics (human baseline: predictability $-2.73$, diversity 94.11); permissive truncation reduces but does not eliminate the mismatch (Appendix~\ref{sec:hyperparameter_details}).

\subsection{Cross-Model Generalization}
\label{sec:cross_model}

So far, classifiers were trained per (model, strategy, configuration) cell. We now ask whether detection generalizes when the configuration is unknown and the generating model is unseen at training time, using the six open models. Classifiers here train on data pooled across all 53 configurations per model, so the same test set can be reused across training regimes; pooling adds within-class variance, so absolute AUC is lower (in-distribution baseline 0.950 vs.\ per-cell 0.974); the relevant statistic is the difference between matched conditions on the same test set.

\paragraph{Matched training-set comparison.} For each model $M$ we evaluate three conditions on the same held-out test set: \textbf{ID-single} trains on $M$ alone; \textbf{ID-pooled} trains on all six models including $M$; and \textbf{LOMO} trains on all models except $M$, so $M$ is unseen. ID-pooled and LOMO are matched in training-set size, so $\Delta = \text{AUC}_{\text{ID-pooled}} - \text{AUC}_{\text{LOMO}}$ isolates the contribution of $M$'s identity in training.

Across the six held-out models, leave-one-model-out training yields mean RF AUC-ROC of 0.929, and $\Delta$ is negative for every model (mean $-0.0044$, 95\% CI $[-0.0063, -0.0025]$; paired $t(5) = -5.90$, $p = 0.002$); for NB and LR, $\Delta$ is indistinguishable from zero (Appendix~\ref{sec:cross_model_details}, Table~\ref{tab:matched_comparison}). Including the held-out model does not improve mean RF performance in this comparison; the NB and LR intervals include zero. Transfer indicates shared feature-level regularities across generators, consistent with a decoding-related distributional mismatch. Identifying the relative contributions of selection, exclusion, and repetition to this transfer requires further analysis.

\paragraph{GPT-2 $\to$ modern transfer.} A complementary test asks whether the signal learned from the oldest, smallest model transfers to far more capable ones. We trained a classifier only on GPT2-XL outputs and evaluated it on each modern open model, alongside a target-specific in-distribution baseline on the same held-out test set.

The classifier trained only on GPT2-XL achieves mean RF AUC 0.949 across the modern targets, within 0.002 of the target-specific baselines (mean $\Delta = +0.0015$, 95\% CI $[-0.0022, +0.0051]$; Appendix~\ref{sec:cross_model_details}); on Llama it reaches 0.987 against an in-distribution 0.991. A classifier trained only on 2019-era outputs thus detects substantially more capable models at near in-distribution accuracy, consistent with the structural account.

\paragraph{Proprietary systems.}\label{sec:proprietary_main} 
GPT-3.5-turbo and Claude-3-Haiku, both proprietary and extensively aligned, remain highly detectable, with AUC-ROC above 0.98 for all three classifiers (Appendix~\ref{sec:proprietary_details}).

\subsection{The Detectability--Quality Dissociation (RQ4)}
\label{sec:quality_dissociation}

The two detection features do not measure contextual appropriateness: high diversity can occur in incoherent text as well as in human writing. Table~\ref{tab:case_study_main} (Appendix~\ref{app:case_study_main}) illustrates repetitive and incoherent machine continuations. Both examples receive high estimated machine probabilities.

We quantify the dissociation with public human ratings from \citet{garces-arias-etal-2025-statistical} (Likert 1--5, $n=300$). The correlation between quality and estimated machine probability is small and non-significant (Spearman $\rho = 0.097$, $p = 0.092$). Beam search, the most detectable strategy, receives the lowest mean rating (2.00), while contrastive search, whose configurations include the least detectable in our grid (AUC-ROC as low as 0.53; Appendix~\ref{sec:extreme_configs}), is rated 2.30 against a human reference of 3.26 (Figure~\ref{fig:quality}, Appendix~\ref{sec:appendix_quality}).

The detectors we study are therefore best understood as detectors of likelihood-constrained generation, not of machine text in general; low classifier confidence does not imply improved quality. Practical recommendations are collected in Appendix~\ref{app:practical}.

\section{Conclusion}
\label{sec:conclusion}

Likelihood-based truncation creates a blind spot between the continuations a model can represent and those its decoder can produce. Our measurements show that human writers regularly select tokens outside common truncation boundaries, with disproportionate exclusion of content-word tokens. Across a large generation benchmark, predictability and lexical diversity provide strong detection signals that transfer across generators, while decoding choices substantially affect detectability. Probability-floor criteria recover more of the human-used tail, pointing to opportunities for improving token selection. These findings motivate evaluating next-token prediction and decoding together: a model can assign positive probability to a human choice while its decoder makes that choice unreachable. The exclusion rate provides a concrete diagnostic for this loss, complementing evaluations of generation quality. Preserving human-like variation requires retaining contextually appropriate choices without sacrificing coherence; lower detectability alone is not sufficient evidence of better generation. Addressing this challenge is an important direction for decoding research.

\section*{Limitations}
(1) Predictability and diversity proxy distributional mismatch but not contextual appropriateness; text can achieve human-like statistics while remaining incoherent. (2) We observe only what writers produced, not why; claims about the communicative function of excluded tokens are interpretive. The POS analysis covers a separate 600-pair subset and three GPT-2 checkpoints, uses automatic tags and subword units, and excludes ambiguous alignments. Its category-prevalence analysis does not control for individual word frequency. Reference-model agreement and within-tokenizer measurement provide robustness checks but do not rule out all modeling or tokenization effects. (3) Scope is constrained to English text from three domains (fiction, encyclopedia, news) at $\leq 256$ tokens; patterns may differ across languages, genres, or longer contexts, and the binary distinction simplifies post-edited scenarios. Since BookCorpus and WikiText likely appear in pretraining corpora, the 8--18\% exclusion may understate the true magnitude; the post-cutoff WikiNews passage in Figure~\ref{fig:decoding_behavior} exhibits comparable exclusion (Section~\ref{sec:truncation_evidence}), but one passage does not establish a lower bound for unseen text. (4) Exclusion analyses cover the five open-source models of Table~\ref{tab:exclusion_main}; the tail-criterion detectability test (Section~\ref{sec:oos}) covers two models and our two features, without quality evaluation; results may vary with other architectures. (5) The per-configuration detection grid splits individual texts, so its test performance does not establish generalization to unseen prompts. (6) For proprietary systems, decoding parameters cannot be controlled beyond API defaults, so the effect of alignment is not identified. The main detection grid also uses inherited generation defaults; nominal settings do not isolate a single truncation rule.

These limitations motivate the natural next steps: extending the out-of-sample detectability experiments of Section~\ref{sec:oos} to all models and strategies with human quality evaluation; benchmarking further adaptive rules such as min-$p$ \citep{nguyen2024minp}, Mirostat \citep{basu2021mirostat}, and adaptive contrastive search \citep{garces-arias-etal-2024-adaptive} on the exclusion diagnostic; and developing adaptive truncation that relaxes bounds at content-bearing positions. Exclusion measures the reachability of human choices, not generation quality, and rules may rank differently under quality-based criteria.

\section*{Ethics Statement}

We use publicly available text datasets, including encyclopedic and news material that can mention identifiable people. Understanding detectability could inform attempts to evade detection; we judge the scientific value of understanding why machine text is detectable to outweigh this risk, particularly as our findings point toward improving generation rather than evading detection. Code and supporting data are available at \url{https://github.com/EstebanGarces/human_vs_machine}. We declare no conflicts of interest.

\section*{Acknowledgments}
Matthias A{\ss}enmacher received funding from the Deutsche Forschungsgemeinschaft (DFG, German Research Foundation) as part of BERD@NFDI, under Grant No.460037581. Esteban Garcés Arias and Matthias A{\ss}enmacher acknowledge the Munich Center for Machine Learning (MCML) for providing financial support and computational resources that enabled the experimental component of this work.

\bibliography{references}

\appendix
\raggedbottom
\section{Appendix}
\label{sec:appendix}

\subsection{Model Checkpoints}
\label{app:checkpoints}

The five open models used in both exclusion analyses (Tables~\ref{tab:exclusion_main} and~\ref{tab:tail_samplers}) correspond to the checkpoints \texttt{gpt2-xl}, \texttt{Qwen/Qwen2-7B}, \texttt{mistralai/Mistral-7B-v0.3}, \texttt{meta-llama/Meta-Llama-3-8B}, and \texttt{tiiuae/falcon-11B}.

\subsection{Detection by Decoding Strategy}
\label{app:strategy_table}

Table~\ref{tab:strategy_performance} reports detection performance by decoding strategy, referenced from Section~\ref{sec:model_results}.

\begin{table}[H]
\centering
\resizebox{\columnwidth}{!}{%
\begin{tabular}{lcccccc}
\toprule
\textbf{Strategy} & \multicolumn{2}{c}{\textbf{RF}} & \multicolumn{2}{c}{\textbf{LR}} & \multicolumn{2}{c}{\textbf{NB}}\\
 & AUC & F1 & AUC & F1 & AUC & F1\\
\midrule
Beam & 0.997 & 0.991 & 0.996 & 0.990 & 0.997 & 0.989 \\
CS & 0.981 & 0.962 & 0.970 & 0.954 & 0.977 & 0.954 \\
Temperature & 0.957 & 0.934 & 0.958 & 0.938 & 0.959 & 0.938 \\
Top-$p$ & 0.956 & 0.918 & 0.958 & 0.921 & 0.957 & 0.919 \\
Top-$k$ & 0.948 & 0.913 & 0.951 & 0.918 & 0.951 & 0.917 \\
\bottomrule
\end{tabular}%
}
\caption{Detection performance by decoding strategy, ordered by mean RF AUC. Means weight model/domain/configuration cells equally within each strategy.}
\label{tab:strategy_performance}
\end{table}

\subsection{Out-of-Sample Detectability}
\label{app:oos_table}

Table~\ref{tab:oos_detectability} lists the per-configuration $\Delta$AUC values underlying Figure~\ref{fig:auc_vs_exclusion} and the analysis of Section~\ref{sec:oos}.

\begin{table}[H]
\centering
\scriptsize
\setlength{\tabcolsep}{2pt}
\begin{tabular}{lrr}
\toprule
Rule, parameter & $\mathcal{R}_{\rm excl}$ (\%) & $\Delta$AUC (RF/LR) \\
\midrule
\multicolumn{3}{l}{\textbf{GPT2-XL}} \\
$\epsilon$, $3{\times}10^{-4}$ & 6.1 & $+0.011$\,/\,$-0.105$ \\
$\eta$, $9{\times}10^{-4}$     & 7.0 & $-0.030$\,/\,$-0.061$ \\
$\epsilon$, $9{\times}10^{-4}$ & 9.7 & $+0.075$\,/\,$+0.078$ \\
$\eta$, $2{\times}10^{-3}$     & 8.9 & $+0.028$\,/\,$+0.026$ \\
typical, $\tau{=}0.9$                   & 9.2 & $-0.006$\,/\,$-0.006$ \\
nucleus, $p{=}0.9$           & 9.2 & \phantom{$+$}0\,/\,\phantom{$+$}0 \\
\midrule
\multicolumn{3}{l}{\textbf{Qwen2-7B}} \\
$\epsilon$, $3{\times}10^{-4}$ & 4.1 & $-0.139$\,/\,$-0.163$ \\
$\eta$, $9{\times}10^{-4}$     & 5.3 & $-0.113$\,/\,$-0.103$ \\
$\epsilon$, $9{\times}10^{-4}$ & 6.5 & $-0.066$\,/\,$-0.051$ \\
$\eta$, $2{\times}10^{-3}$     & 6.9 & $-0.046$\,/\,$-0.039$ \\
typical, $\tau{=}0.9$                   & 9.2 & $-0.002$\,/\,$-0.003$ \\
nucleus, $p{=}0.9$           & 9.2 & \phantom{$+$}0\,/\,\phantom{$+$}0 \\
\bottomrule
\end{tabular}
\caption{Out-of-sample detectability of tail-recovering rules: change in AUC-ROC relative to each model's within-harness nucleus baseline (RF/LR per cell; NB parallels these), prompt-level splits, pooled over domains, each rule applied in isolation; $\mathcal{R}_{\text{excl}}$ is the per-model exclusion rate. Negative values indicate reduced detectability. Human reference features in this experiment are Pred $-2.70$ and Div 94.18, compared with $-2.73$ and 94.11 in the detection grid (Table~\ref{tab:performance}). The nominal $p{=}0.9$ exclusion appears as 8.0\% in Table~\ref{tab:exclusion_main} (detection grid), 8.8\% in Table~\ref{tab:tail_samplers} (five-model tail-criterion average), and 9.2\% here (per-model values).}
\label{tab:oos_detectability}
\end{table}

\subsection{Cross-Generator Transfer}
\label{app:gpt2_transfer}

Table~\ref{tab:gpt2_transfer} reports the GPT-2 $\to$ modern transfer results of Section~\ref{sec:cross_model}.

\begin{table}[H]
\setlength{\tabcolsep}{5pt}
\centering
\small
\begin{tabular}{lccc}
\toprule
\textbf{Target} & \textbf{ID baseline} & \textbf{GPT-2 only} & $\boldsymbol{\Delta}$ \\
\midrule
Llama-3.1 (8B)   & 0.991 & 0.987 & $+0.004$ \\
Mistral-7B & 0.956 & 0.954 & $+0.002$ \\
Qwen 2 (7B)    & 0.940 & 0.943 & $-0.003$ \\
Deepseek (7B)  & 0.956 & 0.952 & $+0.004$ \\
Falcon 2 (11B) & 0.907 & 0.908 & $-0.001$ \\
\midrule
\textbf{Mean} & \textbf{0.950} & \textbf{0.949} & $\boldsymbol{+0.001}$ \\
\bottomrule
\end{tabular}
\caption{GPT-2 $\to$ modern transfer (RF AUC-ROC). A classifier trained only on GPT2-XL outputs is evaluated on each modern open model alongside a target-specific in-distribution baseline.}
\label{tab:gpt2_transfer}
\end{table}

\subsection{Matched Training-Set Comparison}
\label{app:matched_comparison}

Table~\ref{tab:matched_comparison} reports the matched training-set comparison of Section~\ref{sec:cross_model}.

\begin{table}[H]
\centering
\small
\resizebox{\columnwidth}{!}{%
\begin{tabular}{lcccc}
\toprule
\textbf{Held-out $M$} & \textbf{ID-single} & \textbf{ID-pooled} & \textbf{LOMO} & $\boldsymbol{\Delta}$ \\
\midrule
Deepseek  & 0.956 & 0.927 & 0.929 & $-0.001$ \\
Falcon 2  & 0.907 & 0.887 & 0.893 & $-0.006$ \\
GPT2-XL   & 0.952 & 0.925 & 0.929 & $-0.005$ \\
Llama-3.1   & 0.991 & 0.963 & 0.966 & $-0.003$ \\
Mistral-7B & 0.956 & 0.929 & 0.935 & $-0.006$ \\
Qwen 2    & 0.940 & 0.917 & 0.922 & $-0.005$ \\
\midrule
\textbf{Mean} & \textbf{0.950} & \textbf{0.925} & \textbf{0.929} & $\boldsymbol{-0.004}$ \\
\bottomrule
\end{tabular}}
\caption{Matched training-set comparison (RF AUC-ROC). Same test set per held-out model; only training composition varies. Per-classifier statistics in Appendix~\ref{sec:cross_model_details}.}
\label{tab:matched_comparison}
\end{table}

\subsection{Case Study: Quality Versus Detectability}
\label{app:case_study_main}

Table~\ref{tab:case_study_main} contrasts continuations of the same prompt, referenced from Section~\ref{sec:quality_dissociation}.

\begin{table*}[t]
\centering
\footnotesize
\renewcommand{\arraystretch}{1.15}
\resizebox{\textwidth}{!}{%
\begin{tabular}{p{0.17\textwidth}p{0.60\textwidth}c}
\toprule
\textbf{Source} & \textbf{Continuation (excerpt)} & $\boldsymbol{P(\text{Machine})}$ \\
\midrule
Llama-3.1,\newline beam (3) & \cellcolor{repetitive}``he served as Commanding Officer of the 1st Battalion, The Queen's Royal Regiment [...] Commanding Officer of the 2nd Battalion, The Queen's Royal Regiment [...] of the 3rd Battalion [...]'' & 1.00 \\
Qwen 2,\newline CS (1.0, 20) & \cellcolor{erratic}``Headelm oversa Operation Agulhene535af69, to reoccupyo portug 1 st colonial outst a nces in AFRica coun trie s to preva l sepa rabli st mov [...]'' & 0.88 \\
\bottomrule
\end{tabular}}
\caption{Continuations of the same WikiText prompt (Headlam's service record), with the Random Forest estimate of machine authorship. Degenerate (repetitive) text is highlighted in \colorbox{repetitive}{purple}; erratic (incoherent) text in \colorbox{erratic}{brown}. Both machine continuations receive high estimated machine probabilities (1.00 and 0.88), despite exhibiting different forms of poor quality. These individual scores are not measures of population-level detection performance or calibrated certainty. Full case study in Table~\ref{tab:case_study_3} (Appendix~\ref{sec:case_study}).}
\label{tab:case_study_main}
\end{table*}

\subsection{Extended Related Work}
\label{app:extended_rw}

The detection of machine-generated text has attracted significant attention as LLM capabilities have advanced \citep{jawahar2020automatic}. Existing approaches fall into three main categories. \textit{Zero-shot statistical methods} exploit properties of model probability distributions: DetectGPT \citep{mitchell2023detectgpt} identifies machine text via negative curvature in the log-probability landscape, while likelihood-based approaches exploit differences in token predictability \citep{gehrmann2019gltr, ippolito2020automatic}. \textit{Trained classifiers} distinguish human from machine text using labeled corpora \citep{solaiman2019release, uchendu2020authorship, tang2024science}, but often struggle to generalize across models and domains. \textit{Watermarking methods} embed statistical signatures during generation \citep{kirchenbauer2023watermark}, though these can be undermined by paraphrasing attacks \citep{krishna2023paraphrasing}. Our analysis connects these statistical detection signals to the tokens removed by likelihood-based truncation.

Decoding strategies strongly shape the properties of generated text \citep{wiher-etal-2022-decoding, garces-arias-etal-2025-decoding, FRISONI2026108249}. Prior work has largely focused on downstream task performance (e.g., summarization), whereas we examine human-token exclusion and detectability across models and domains. \citet{holtzman2020how} show that maximization-based decoding leads to degeneration, motivating nucleus (top-$p$) sampling to truncate the unreliable tail. Earlier work introduced top-$k$ sampling \citep{fan2018hierarchical}, while contrastive search balances likelihood with diversity \citep{su2023contrastivesearchneedneural}. From an information-theoretic perspective, locally typical sampling targets tokens with information content close to conditional entropy \citep{meister2023locally}, drawing on accounts of information density and communicative efficiency \citep{jaeger2010redundancy, gibson2019efficiency}. We extend this line of work by linking truncation directly to detectability: mechanisms that improve fluency by excluding low-probability tokens also introduce systematic deviations from human text.

Several studies bear directly on our analysis. \citet{ippolito2020automatic} show that decoding choices tuned to fool human readers introduce statistical regularities that make automatic detection easier; we account for that trade-off mechanistically, tracing it to the tokens truncation removes. \citet{hewitt2022truncation} cast truncation as desmoothing and propose $\eta$- and $\epsilon$-sampling to recover the support of the true distribution more judiciously; we take the complementary, diagnostic view, asking what such truncation excludes relative to human text, and evaluate their samplers directly against our exclusion benchmark (Section~\ref{sec:truncation_evidence}). \citet{finlayson2024closing} use the softmax bottleneck to derive a truncation criterion and evaluate it in pilot generation experiments. Our diagnostic measures human-token exclusion and its relationship with detectability. Closest to our setting, \citet{dubois-etal-2025-sampling} show that small changes in sampling parameters can sharply alter detector accuracy; our results concur that decoding governs detectability, and add a mechanistic account in terms of token exclusion, a linguistic decomposition of the excluded set, and coverage of proprietary and non-Transformer models.

Related approaches also address the predictive distribution and its evaluation. Variance-Calibrated Modulation reshapes logits before truncation using contextual relevance and a variance-scaled repetition penalty \citep{ding2026breakinglikelihoodtrapvariancecalibrated}. Credal-set analysis measures gaps between model outputs and human creative variation \citep{garces-arias-etal-2025-geometry}, while multicriteria evaluation compares decoding strategies across quality dimensions \citep{garces-arias-etal-2025-towards}. These perspectives complement exclusion rates with distributional and quality-based assessments.

Psycholinguistics and information theory offer complementary accounts of language use. Surprisal theory relates processing difficulty to negative log-probability \citep{hale2001probabilistic, levy2008expectation}, with evidence across eleven languages \citep{wilcox2023testing}. Speech-production models distinguish conceptual preparation, lexical selection, and articulation \citep{levelt1989speaking, indefrey2004spatial}. Information-density research examines how speakers distribute information across an utterance \citep{jaeger2010redundancy}, while \citet{shannon1951prediction} estimates the entropy and redundancy of English. These perspectives motivate studying the information retained by decoding rules.

\subsection{Token Category Taxonomy}
\label{sec:token_taxonomy}

Table~\ref{tab:token_taxonomy_appendix} formalizes the four-category taxonomy introduced in Section~\ref{sec:preliminaries}.

\begin{table}[H]
\centering
\footnotesize
\setlength{\tabcolsep}{3pt}
\renewcommand{\arraystretch}{1.15}
\begin{tabular}{c|>{\centering\arraybackslash}m{0.42\columnwidth}|>{\centering\arraybackslash}m{0.42\columnwidth}}
\toprule
 & \textbf{Human Includes} & \textbf{Human Excludes} \\
\midrule
\rotatebox[origin=c]{90}{\textbf{Model Includes}} &
\cellcolor{cellgreen}\textbf{Shared Inclusion}\par
\cellcolor{cellgreen}\footnotesize $\mathcal{T}_{\mathcal{H}}(\mathbf{x}_{<t}) \cap \mathcal{T}_{\theta}(\mathbf{x}_{<t})$ Tokens that are both statistically probable for the model and communicatively appropriate for humans
&
\cellcolor{cellyellow}\textbf{Model Overreach}\par
\cellcolor{cellyellow}\footnotesize $\mathcal{E}_{\mathcal{H}}(\mathbf{x}_{<t}) \cap \mathcal{T}_{\theta}(\mathbf{x}_{<t})$ Tokens included by the model's strategy but excluded by humans as communicatively inappropriate
\\
\midrule
\rotatebox[origin=c]{90}{\textbf{Model Excludes}} &
\cellcolor{cellred}\textbf{Blind Spot}\par
\cellcolor{cellred}\footnotesize $\mathcal{T}_{\mathcal{H}}(\mathbf{x}_{<t}) \cap \mathcal{E}_{\theta}(\mathbf{x}_{<t})$ \par
\cellcolor{cellred}\footnotesize \textit{(Focus of this paper)} Tokens that are communicatively valuable, yet are excluded by the likelihood-based truncation
&
\cellcolor{cellgray}\textbf{Shared Exclusion}\par
\cellcolor{cellgray}\footnotesize $\mathcal{E}_{\mathcal{H}}(\mathbf{x}_{<t}) \cap \mathcal{E}_{\theta}(\mathbf{x}_{<t})$ Tokens rejected by both systems (e.g., rarely observed in training data, typically due to grammatical or semantic unsuitability)
\\
\bottomrule
\end{tabular}
\caption{Taxonomy of token categories based on model and human selection mechanisms. The \textit{blind spot} contains tokens that humans may select for communicative purposes, but that likelihood-based truncation systematically excludes due to low probability.}
\label{tab:token_taxonomy_appendix}
\end{table}

\subsection{Extended Discussion}
\label{sec:discussion}

\subsubsection{Summary of Evidence}

Our results provide evidence consistent with the hypothesis that the detectability of machine-generated text arises from a distributional mismatch introduced at decoding time, when likelihood-based truncation removes low-rank tokens that human writers use and that the model itself assigns nonzero probability. Under typical truncation settings, 8--18\% of human-selected tokens would be excluded, demonstrating that the blind spot is empirically substantial (\textbf{RQ1}). Machine text exhibits systematically higher predictability than human text, and this predictability serves as the primary signal enabling detection with simple classifiers (\textbf{RQ2}). Stricter truncation increases both the exclusion rate and detectability; relaxing truncation reduces both. Decoding parameters affect detectability, while scale and architecture also show effects in some analyses (\textbf{RQ3}). These patterns are consistent with the mechanistic explanation that likelihood-based truncation creates a systematic blind spot. Finally, low detectability provides limited information about perceived quality (\textbf{RQ4}).

\subsubsection{Why Detection Generalizes Across Models}

Cross-model transfer shows that the two features capture regularities shared by multiple generators. Within-family truncation-set overlap and the GPT-2 POS asymmetry provide complementary evidence about shared candidates and the linguistic composition of excluded tokens. Distinguishing their contributions to transfer is a direction for future work.

\subsubsection{Limitations of Current Approaches}

The weak association between detectability and human quality ratings highlights the need for separate quality evaluation. Current decoding strategies face a trade-off: strict truncation produces coherent but detectable text (restricted to high-probability tokens), while permissive truncation produces less detectable but often incoherent text (including inappropriate low-probability tokens alongside appropriate ones). The core problem is that likelihood-based truncation cannot distinguish between two types of low-probability tokens: those that are contextually appropriate (and would be selected by a human pursuing specific communicative goals) and those that are contextually inappropriate. Permissive truncation admits both; strict truncation excludes both. Addressing this limitation would require mechanisms that can identify contextually appropriate tokens regardless of their probability ranking. Probability-floor criteria are a partial step: they widen the pool where the distribution is flat and recover much of the excluded tail (Section~\ref{sec:truncation_evidence}), but they still select by probability alone and accordingly narrow rather than close the blind spot.

\subsection{Practical Implications}
\label{app:practical}

\paragraph{For detection research.} Predictability and diversity suffice when the decoding configuration is known and degrade gracefully across generators and even across unseen generating models (Section~\ref{sec:cross_model}). Classifier confidence, however, does not track perceived quality (Section~\ref{sec:quality_dissociation}); detection systems built on similar features should not equate low classifier confidence with improved generation quality, and should report calibration separately for coherent and incoherent regimes.

\paragraph{For decoding research.} The limitation is not truncation itself but the truncation criterion: likelihood alone cannot separate rare but apt tokens from rare and unsuitable ones. The exclusion rate, together with its part-of-speech decomposition (Section~\ref{sec:pos_main}), provides a quantitative benchmark for evaluating tail-aware sampling methods; probability-floor criteria already dominate nucleus sampling on this benchmark (Section~\ref{sec:truncation_evidence}).

\paragraph{For alignment.} The evaluated proprietary systems remain detectable, but their training procedures and decoding settings are not independently controlled. These results motivate controlled comparisons of aligned and unaligned checkpoints under identical decoding conditions.

\subsection{Human Token Exclusion Rates}
\label{sec:exclusion_details}

Tables~\ref{tab:exclusion_rates} and~\ref{tab:rank_distribution} quantify the magnitude of the blind spot discussed in Section~\ref{sec:truncation_evidence}. Exclusion is computed entirely within each model's own tokenizer: each text tokenized by that model, ranks taken under its distribution, exclusion judged against its truncation set; ranks are never compared across vocabularies. We average only the resulting per-model rates, which are unitless.

\begin{table}[H]
\setlength{\tabcolsep}{5pt}
\centering
\small
\begin{tabular}{lcccc}
\toprule
\textbf{Truncation} & \multicolumn{3}{c}{\textbf{Exclusion Rate (\%)}} & \textbf{Avg.} \\
\textbf{Setting} & Book & News & Wiki & \textbf{(95\% CI)} \\
\midrule
\multicolumn{5}{l}{\textit{Top-$k$ Sampling}} \\
$k=1$ & 49.5 & 47.8 & 51.3 & 49.5 [47.5, 51.9] \\
$k=5$ & 25.0 & 22.5 & 27.4 & 25.0 [23.1, 27.4] \\
$k=10$ & 18.1 & 15.7 & 20.4 & 18.1 [16.3, 20.2] \\
$k=20$ & 12.9 & 10.8 & 15.2 & 13.0 [11.4, 14.8] \\
$k=50$ & 8.0 & 6.5 & 9.9 & 8.1 [6.9, 9.5] \\
$k=100$ & 5.4 & 4.3 & 7.0 & 5.6 [4.7, 6.7] \\
\midrule
\multicolumn{5}{l}{\textit{Top-$p$ (Nucleus) Sampling}} \\
$p=0.6$ & 28.1 & 28.5 & 29.6 & 28.7 [27.7, 29.8] \\
$p=0.8$ & 15.0 & 16.1 & 15.8 & 15.6 [14.9, 16.5] \\
$p=0.9$ & 7.4 & 8.7 & 7.9 & 8.0 [7.4, 8.8] \\
$p=0.95$ & 3.5 & 4.5 & 3.7 & 3.9 [3.4, 4.6] \\
$p=0.99$ & 0.6 & 0.9 & 0.8 & 0.8 [0.6, 1.0] \\
\bottomrule
\end{tabular}
\caption{Human token exclusion rates $\mathcal{R}_{\text{excl}}$ across truncation configurations. Values are averaged over five language models (GPT2-XL, Qwen2-7B, Mistral-7B-v0.3, Llama-3-8B, Falcon2-11B) and all texts in each dataset split; 95\% confidence intervals are computed over this model $\times$ text distribution. Rates are computed within each model's own tokenizer; ranks are never compared across vocabularies, and only the resulting unitless rates are averaged.}
\label{tab:exclusion_rates}
\end{table}

\begin{table}[H]
\centering
\small
\begin{tabular}{lccc}
\toprule
\textbf{Rank Range} & \textbf{Book} & \textbf{News} & \textbf{Wiki} \\
\midrule
$[1, 5]$ & 75.0 & 77.5 & 72.6 \\
$[6, 10]$ & 6.9 & 6.8 & 7.0 \\
$[11, 20]$ & 5.2 & 4.9 & 5.3 \\
$[21, 50]$ & 4.9 & 4.3 & 5.2 \\
$[51, 100]$ & 2.6 & 2.2 & 2.9 \\
$> 100$ & 5.4 & 4.3 & 7.0 \\
\midrule
Median Rank & 1.6 & 1.2 & 1.8 \\
Mean Rank & 75.5 & 50.5 & 155.0 \\
\bottomrule
\end{tabular}
\caption{Distribution of human token ranks according to language model probability distributions, averaged across five models (GPT2-XL, Qwen2-7B, Mistral-7B-v0.3, Llama-3-8B, Falcon2-11B) and all texts in each dataset.}
\label{tab:rank_distribution}
\end{table}

\subsection{Detection Performance by Classifier and Model}
\label{sec:model_details}

Table~\ref{tab:overall_performance} reports aggregate classification performance across the full experimental grid, referenced from Section~\ref{sec:predictability_results}; Tables~\ref{tab:model_performance} and~\ref{tab:model_size} break performance down by generating model.

\begin{table}[H]
\centering
\small
\begin{tabular}{lcccc}
\toprule
\textbf{Classifier} & \multicolumn{2}{c}{\textbf{AUC-ROC}} & \multicolumn{2}{c}{\textbf{F1-Score}} \\
 & Mean & Std & Mean & Std \\
\midrule
RF & 0.974 & 0.054 & 0.953 & 0.071 \\
LR & 0.968 & 0.068 & 0.949 & 0.077 \\
NB & 0.972 & 0.054 & 0.949 & 0.069 \\
\bottomrule
\end{tabular}
\caption{Classification performance across all configurations ($n=1{,}062$ experiments), averaged over all models, datasets, and decoding configurations.}
\label{tab:overall_performance}
\end{table}

\begin{table}[H]
\scriptsize
\centering
\resizebox{\columnwidth}{!}{%
\begin{tabular}{lccc}
\toprule
\textbf{Model} & \textbf{RF AUC} & \textbf{LR AUC} & \textbf{NB AUC}\\
\midrule
Llama-3.1 & 0.993 $\pm$ 0.007 & 0.994 $\pm$ 0.006 & 0.993 $\pm$ 0.007 \\
GPT2-XL & 0.980 $\pm$ 0.022 & 0.979 $\pm$ 0.024 & 0.978 $\pm$ 0.026 \\
Mistral-7B & 0.980 $\pm$ 0.025 & 0.981 $\pm$ 0.024 & 0.979 $\pm$ 0.028 \\
Deepseek & 0.976 $\pm$ 0.042 & 0.952 $\pm$ 0.069 & 0.968 $\pm$ 0.047 \\
Qwen 2 & 0.969 $\pm$ 0.035 & 0.964 $\pm$ 0.039 & 0.967 $\pm$ 0.033 \\
Falcon 2 & 0.945 $\pm$ 0.041 & 0.938 $\pm$ 0.050 & 0.948 $\pm$ 0.044 \\
\bottomrule
\end{tabular}%
}
\caption{Average detectability by model (ordered by RF AUC-ROC). GPT2-XL is more detectable than some recent models, while Llama-3.1 has the highest mean AUC in this grid.}
\label{tab:model_performance}
\end{table}

\begin{table}[H]
\centering
\resizebox{\columnwidth}{!}{
\begin{tabular}{lcc}
\toprule
\textbf{Model} & \textbf{Parameters} & \textbf{Avg. AUC-ROC} \\
\midrule
GPT2-XL & 1.5B & 0.979 \\
Falcon 2 & 11B & 0.944 \\
Qwen 2 & 7B & 0.967 \\
Mistral-7B & 7B & 0.980 \\
Deepseek & 7B & 0.965 \\
Llama-3.1 & 8B & 0.993 \\
\bottomrule
\end{tabular}
}
\caption{Model size versus detectability. No clear correlation emerges between parameter count and detection rates.}
\label{tab:model_size}
\end{table}

\subsection{Cross-Model Generalization: Statistical Details}
\label{sec:cross_model_details}

This appendix reports per-classifier statistics for the matched training-set comparison and the GPT-2 $\to$ modern transfer experiments described in Section~\ref{sec:cross_model}.

\paragraph{Setup details.} For both experiments, classifiers are trained on data pooled across all decoding configurations (rather than per-cell as in Section~\ref{sec:classification}), to enable cross-model comparison on the same test set. The 80/20 train/test split is performed at the prompt level (rather than the row level) to prevent leakage. RF uses 500 trees with $\texttt{class\_weight=balanced}$; LR uses $\texttt{max\_iter=2000}$ with $\texttt{class\_weight=balanced}$; NB is Gaussian. All experiments use the same random seed and the same global train/test prompt partition.

\paragraph{Matched comparison: per-classifier statistics.} The quantity reported is $\Delta = \text{AUC}_{\text{ID-pooled}} - \text{AUC}_{\text{LOMO}}$, averaged across the six held-out models. 95\% confidence intervals are computed from the per-model $\Delta$ values using the $t$-distribution.

\begin{table}[H]
\centering
\small
\resizebox{\columnwidth}{!}{%
\begin{tabular}{lccccc}
\toprule
\textbf{Classifier} & \textbf{Mean $\Delta$} & \textbf{95\% CI} & \textbf{Sign test} & \textbf{Paired $t(5)$} & \textbf{$p$} \\
\midrule
RF & $-0.0044$ & $[-0.0063, -0.0025]$ & $p = 0.031$ (6/6 negative) & $t = -5.90$ & $0.002$ \\
NB & $-0.00002$ & $[-0.0002, +0.0002]$ & $p = 0.69$ (2/6 negative) & $t = -0.21$ & $0.84$ \\
LR & $-0.0024$ & $[-0.0195, +0.0147]$ & $p = 1.00$ (3/6 negative) & $t = -0.36$ & $0.73$ \\
\bottomrule
\end{tabular}}
\caption{Per-classifier statistics for the matched cross-model comparison. Across all three classifiers, the 95\% CI for $\Delta$ excludes any positive value larger than 0.015; for RF the entire interval is negative, while the NB interval spans zero.}
\label{tab:matched_stats}
\end{table}

\paragraph{GPT-2 $\to$ modern transfer: per-classifier statistics.} The quantity reported is $\Delta = \text{AUC}_{\text{ID-target}} - \text{AUC}_{\text{GPT-2}\to\text{target}}$, averaged across the five modern open targets.

\begin{table}[H]
\centering
\small
\resizebox{\columnwidth}{!}{%
\begin{tabular}{lcccc}
\toprule
\textbf{Classifier} & \textbf{Mean $\Delta$} & \textbf{95\% CI} & \textbf{Paired $t(4)$} & \textbf{$p$} \\
\midrule
RF & $+0.0015$ & $[-0.0022, +0.0051]$ & $t = +1.12$ & $0.33$ \\
NB & $+0.0005$ & $[-0.0013, +0.0023]$ & $t = +0.75$ & $0.49$ \\
LR & $+0.0090$ & $[-0.0233, +0.0412]$ & $t = +0.77$ & $0.48$ \\
\bottomrule
\end{tabular}}
\caption{Per-classifier statistics for the GPT-2 $\to$ modern transfer. The difference between target-specific in-distribution training and GPT-2-only training is not statistically detectable for any classifier.}
\label{tab:gpt2_stats}
\end{table}

\paragraph{Interpretation.} Two interpretive points are worth emphasizing. First, absolute AUC values in the pooled-training setting are lower than the per-cell AUCs reported in Section~\ref{sec:predictability_results} (mean RF ID baseline $0.950$ vs.\ per-cell mean $0.974$). This is expected: pooling across decoding configurations introduces within-class variance from the full hyperparameter sweep, which a per-cell classifier never sees. The relevant quantity for assessing cross-model generalization is the \emph{difference} between matched conditions on the same test set, not the absolute level. Second, the LR confidence interval is wider than RF and NB because LR's linear decision boundary is suboptimal for the pooled feature distribution; this manifests in higher per-cell variance in $\Delta$, not in a systematic deviation from zero.

\subsection{Class Balance Robustness}
\label{sec:class_balance}

Because the corpus contains $\sim$1.8M machine-generated texts and 5{,}261 human references, we assessed sensitivity to class imbalance. We repeated the matched cross-model comparison with machine examples downsampled to the human count (5,261 each) and observed essentially unchanged AUC-ROC: mean changes were $\leq 0.001$ for NB and RF, and $\leq 0.018$ for LR (the slightly larger LR shift is itself within the within-classifier variance reported in Table~\ref{tab:matched_stats}). This is consistent with AUC-ROC's invariance to class prior and indicates that the imbalance does not drive the reported separability.

\subsection{Domain-Specific Patterns}
\label{sec:domain_details}

\begin{table}[H]
\centering
\resizebox{\columnwidth}{!}{%
\begin{tabular}{lccc}
\toprule
\textbf{Dataset} & \textbf{RF AUC} & \textbf{LR AUC} & \textbf{NB AUC}\\
\midrule
BookCorpus & 0.982 $\pm$ 0.030 & 0.976 $\pm$ 0.039 & 0.979 $\pm$ 0.034\\
WikiNews & 0.960 $\pm$ 0.043 & 0.951 $\pm$ 0.055 & 0.960 $\pm$ 0.047\\
WikiText & 0.980 $\pm$ 0.023 & 0.976 $\pm$ 0.028 & 0.977 $\pm$ 0.027\\
\bottomrule
\end{tabular}%
}
\caption{Detection performance by dataset domain. WikiNews shows the lowest average detectability.}
\label{tab:dataset_performance}
\end{table}

\begin{table}[H]
\centering
\small
\begin{tabular}{lcccc}
\toprule
\textbf{Dataset} & \multicolumn{2}{c}{\textbf{Predictability}} & \multicolumn{2}{c}{\textbf{Diversity}} \\
 & Mean & Var & Mean & Var \\
\midrule
BookCorpus & $-2.94$ & 0.104 & 94.98 & 25.54 \\
WikiNews & $-2.41$ & 0.077 & 93.84 & 18.18 \\
WikiText & $-2.83$ & 0.155 & 93.50 & 34.49 \\
\bottomrule
\end{tabular}
\caption{Human text statistics across domains.}
\label{tab:feature_stats}
\end{table}

WikiNews shows both the lowest detectability and the highest human predictability ($-2.41$), suggesting that formulaic, factual writing is inherently more predictable and thus easier for likelihood-based generation to approximate. BookCorpus shows the highest detectability, consistent with creative writing requiring more tokens from outside the high-probability region. The inverse relationship between human predictability and detectability across domains supports the framework: domains where human text already favors high-probability tokens leave less room for the distributional mismatch that enables detection.

\subsection{Extreme Configurations}
\label{sec:extreme_configs}

Tables~\ref{tab:top_distinguishable} and~\ref{tab:least_distinguishable} list the most and least detectable configurations in the grid.

\begin{table}[H]
\centering
\small
\resizebox{\columnwidth}{!}{%
\begin{tabular}{llllcc}
\toprule
Model & Strategy & Config. & Domain & Classifier & AUC \\
\midrule
Deepseek & CS & (1.0, 15) & Book & LR & 1.000 \\
Deepseek & CS & (1.0, 15) & Book & RF & 1.000 \\
Deepseek & CS & (1.0, 50) & Book & LR & 1.000 \\
Deepseek & CS & (1.0, 50) & Book & RF & 1.000 \\
Deepseek & beam & 15 & Book & LR & 1.000 \\
Deepseek & beam & 15 & Book & NB & 1.000 \\
Deepseek & beam & 15 & Book & RF & 1.000 \\
Deepseek & beam & 20 & Book & LR & 1.000 \\
Deepseek & beam & 20 & Book & RF & 1.000 \\
\bottomrule
\end{tabular}}
\caption{Selected evaluations achieving the highest recorded AUC in the grid; contrastive-search examples use $k\geq3$. Many evaluations tie at the stored precision; rows are examples from that tied set, not a unique ranking.}
\label{tab:top_distinguishable}
\end{table}

\begin{table}[H]
\centering
\small
\resizebox{\columnwidth}{!}{%
\begin{tabular}{llllcc}
\toprule
Model & Strategy & Config. & Domain & Classifier & AUC \\
\midrule
Deepseek & CS & (0.6, 20) & Book & LR & 0.529 \\
Falcon 2 & temp & 1 & News & LR & 0.532 \\
Qwen 2 & CS & (0.8, 3) & Wiki & LR & 0.537 \\
Deepseek & CS & (0.6, 15) & Wiki & LR & 0.569 \\
Qwen 2 & CS & (0.8, 3) & Wiki & NB & 0.570 \\
Falcon 2 & CS & (0.8, 3) & News & LR & 0.608 \\
Deepseek & CS & (0.6, 3) & News & LR & 0.610 \\
GPT2-XL & temp & 1 & News & LR & 0.615 \\
Deepseek & CS & (0.8, 5) & Book & LR & 0.616 \\
GPT2-XL & temp & 1 & News & RF & 0.617 \\
\bottomrule
\end{tabular}}
\caption{Ten lowest-AUC classifier/configuration evaluations in the 59-configuration grid. A configuration can occur for more than one classifier.}
\label{tab:least_distinguishable}
\end{table}

Low-AUC configurations occur for several models and include contrastive search and temperature sampling. These examples show the importance of reporting the decoding setting and classifier alongside the model.

\subsection{Proprietary Models}
\label{sec:proprietary_details}

\begin{figure*}[tp]
\centering
  \includegraphics[width=0.92\textwidth]{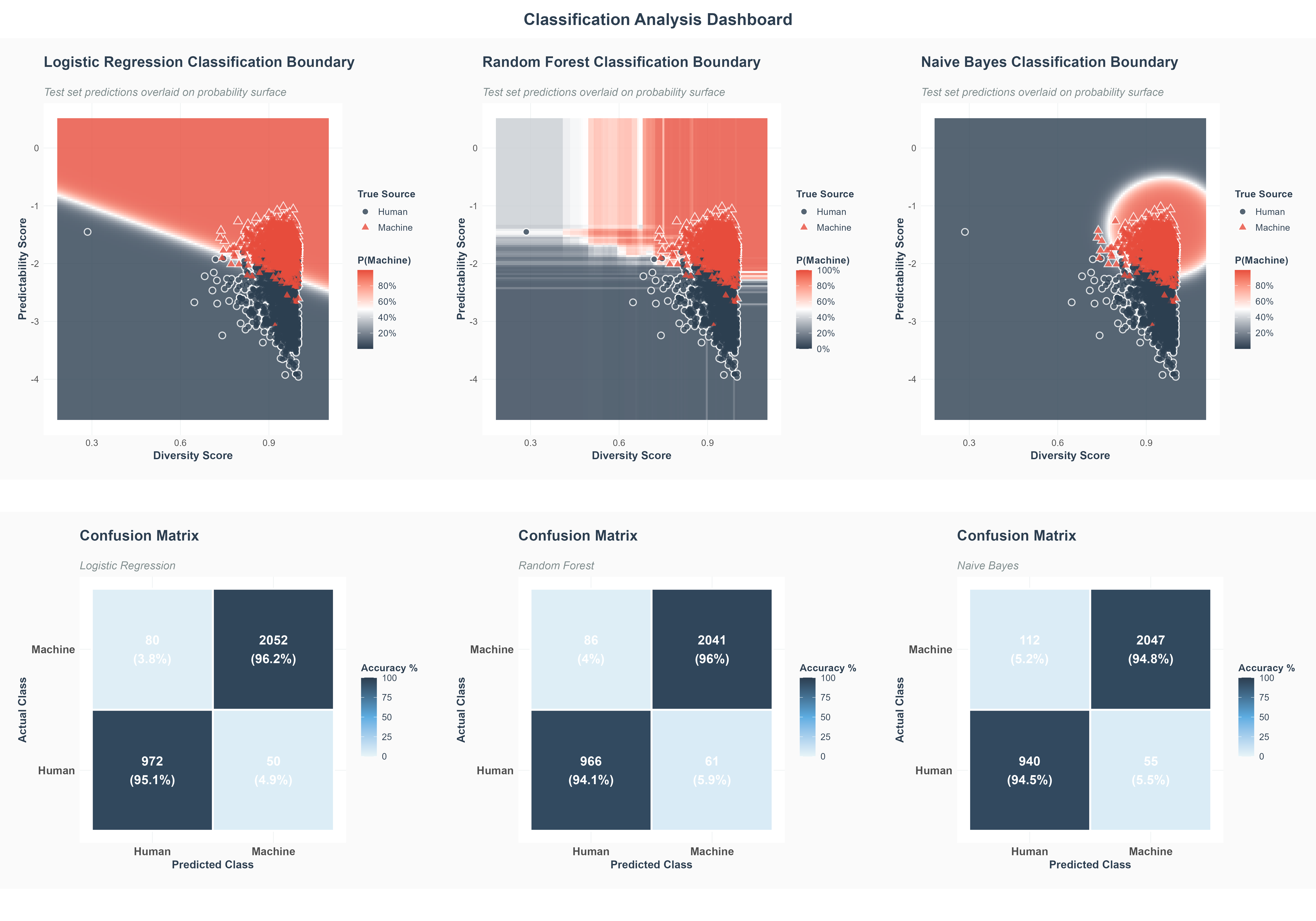}
  \caption{Binary classification performance for proprietary models GPT-3.5-turbo and Claude-3-Haiku. Despite advanced training procedures and alignment, both remain highly detectable using only predictability and diversity features.}
  \label{fig:sota_performance}
\end{figure*}

Figure~\ref{fig:sota_performance} and Table~\ref{tab:sota_performance_metrics} present detection results for GPT-3.5-turbo and Claude-3-Haiku, referenced from Section~\ref{sec:cross_model}.

\begin{table*}[tp]
\centering
\resizebox{0.62\textwidth}{!}{%
\begin{tabular}{@{}l*{7}{c}@{}}
\toprule
\multirow{2}{*}{\textbf{Classifier}} & \multicolumn{7}{c}{\textbf{Performance Metrics}} \\
\cmidrule(l){2-8}
 & Acc. & Prec. & Rec. & F1 & Spec. & AUC-ROC & AUC-PR \\
\midrule
RF & 0.953 & 0.960 & 0.971 & 0.965 & 0.918 & 0.983 & 0.988 \\
LR & \textbf{0.959} & \textbf{0.963} & \textbf{0.976} & \textbf{0.969} & \textbf{0.924} & \textbf{0.989} & \textbf{0.994} \\
NB & 0.947 & 0.948 & 0.974 & 0.961 & 0.894 & 0.988 & 0.993 \\
\bottomrule
\end{tabular}}
\caption{Detection performance for proprietary models (GPT-3.5-turbo and Claude-3-Haiku combined).}
\label{tab:sota_performance_metrics}
\end{table*}

The logistic regression coefficients quantify feature contributions:
\begin{equation}
\begin{aligned}
\log\left(\frac{P(\text{Human})}{P(\text{Machine})}\right)
&= 6.24 + 23.19 \cdot \text{Div} \\
&\quad - 12.88 \cdot \text{Pred}.
\end{aligned}
\end{equation}
Both coefficients are significant ($p < 2 \times 10^{-16}$). Higher lexical diversity is associated with human classification; higher predictability is associated with machine classification, consistent with the prediction that likelihood-based truncation concentrates token selection in high-probability regions.

GPT-3.5-turbo and Claude-3-Haiku remain highly detectable in this evaluation (AUC-ROC $>0.98$). Because the study does not compare aligned and unaligned versions under identical decoding conditions, these observations do not identify the effect of alignment on detectability.

\subsection{Full Hyperparameter Results}
\label{sec:hyperparameter_details}

Table~\ref{tab:performance} presents detection performance and text properties for 48 non-temperature configurations: beam search, top-$k$, top-$p$, and contrastive search with $k\geq3$.

\begin{table*}[p]
\centering
\small
\resizebox{0.88\textwidth}{!}{%
\begin{tabular}{@{}llcccccc@{}}
\toprule
\multirow{2}{*}{\textbf{Strategy}} & \multirow{2}{*}{\textbf{Config.}} & \multicolumn{4}{c}{\textbf{Classification Metrics}} & \multicolumn{2}{c}{\textbf{Text Properties}} \\
\cmidrule(lr){3-6} \cmidrule(lr){7-8}
 & & Prec. & Rec. & F1 & AUC-ROC & Pred. & Div. \\
\midrule
\multirow{6}{*}{\rotatebox[origin=c]{90}{\scriptsize Beam Search}}
& 3 & 0.987 & 0.993 & \cellcolor{overallbest}0.989 & 0.996 & $-1.67 \pm 1.15$ & \cellcolor{stratbest}$59.20 \pm 40.76$ \\
& 5 & \cellcolor{overallbest}0.986 & 0.993 & 0.990 & \cellcolor{overallbest}0.995 & $-1.68 \pm 1.14$ & $58.72 \pm 41.08$ \\
& 10 & 0.987 & 0.994 & 0.991 & 0.996 & $-1.67 \pm 1.14$ & $57.57 \pm 41.74$ \\
& 15 & 0.987 & \cellcolor{overallbest}0.992 & 0.990 & 0.996 & $-1.67 \pm 1.15$ & $56.91 \pm 42.14$ \\
& 20 & 0.989 & 0.994 & 0.992 & 0.997 & \cellcolor{stratbest}$-1.69 \pm 1.13$ & $56.40 \pm 42.43$ \\
& 50 & 0.989 & 0.995 & 0.992 & 0.997 & $-1.66 \pm 1.15$ & $54.91 \pm 43.41$ \\
\midrule
\multirow{30}{*}{\rotatebox[origin=c]{90}{\scriptsize Contrastive Search}}
& (0.2, 3) & 0.968 & 0.985 & 0.976 & 0.989 & $-1.89 \pm 1.00$ & $64.18 \pm 37.60$ \\
& (0.2, 5) & 0.973 & 0.985 & 0.979 & 0.991 & $-1.82 \pm 1.06$ & $65.38 \pm 36.84$ \\
& (0.2, 10) & 0.970 & 0.982 & 0.976 & 0.990 & $-1.87 \pm 1.01$ & $66.72 \pm 36.06$ \\
& (0.2, 15) & 0.966 & 0.981 & 0.973 & 0.988 & $-1.87 \pm 0.98$ & $67.51 \pm 35.69$ \\
& (0.2, 20) & 0.972 & 0.987 & 0.979 & 0.990 & $-1.84 \pm 1.03$ & $68.35 \pm 35.34$ \\
& (0.2, 50) & 0.963 & 0.980 & 0.971 & 0.985 & $-1.87 \pm 1.01$ & $70.97 \pm 34.03$ \\
& (0.4, 3) & 0.960 & 0.972 & 0.966 & 0.985 & $-2.17 \pm 0.88$ & $73.71 \pm 30.71$ \\
& (0.4, 5) & 0.941 & 0.963 & 0.951 & 0.973 & $-2.10 \pm 0.94$ & $76.79 \pm 28.30$ \\
& (0.4, 10) & 0.930 & 0.956 & 0.943 & 0.966 & $-2.22 \pm 0.89$ & $80.06 \pm 25.88$ \\
& (0.4, 15) & 0.941 & 0.967 & 0.954 & 0.974 & $-1.99 \pm 0.90$ & $82.33 \pm 24.17$ \\
& (0.4, 20) & 0.954 & 0.965 & 0.959 & 0.976 & $-2.15 \pm 0.93$ & $83.82 \pm 22.83$ \\
& (0.4, 50) & 0.948 & 0.963 & 0.955 & 0.969 & $-2.06 \pm 0.87$ & $87.11 \pm 19.29$ \\
& (0.6, 3) & \cellcolor{overallbest}\textbf{0.852}$\star$ & \cellcolor{overallbest}0.891 & \cellcolor{overallbest}\textbf{0.870}$\star$ & \cellcolor{overallbest}\textbf{0.901}$\star$ & $-2.27 \pm 0.78$ & $81.67 \pm 23.77$ \\
& (0.6, 5) & 0.923 & 0.943 & 0.933 & 0.961 & $-2.26 \pm 0.80$ & $85.05 \pm 20.83$ \\
& (0.6, 10) & 0.886 & 0.906 & 0.895 & 0.934 & $-2.38 \pm 0.76$ & $87.84 \pm 18.19$ \\
& (0.6, 15) & 0.898 & 0.911 & 0.904 & 0.934 & $-2.25 \pm 0.81$ & $88.88 \pm 17.35$ \\
& (0.6, 20) & 0.920 & 0.938 & 0.929 & 0.949 & $-2.33 \pm 0.83$ & $89.42 \pm 16.97$ \\
& (0.6, 50) & 0.934 & 0.933 & 0.933 & 0.962 & $-2.45 \pm 0.85$ & $90.85 \pm 16.00$ \\
& (0.8, 3) & 0.883 & 0.906 & 0.893 & 0.916 & $-2.90 \pm 1.01$ & $87.39 \pm 18.49$ \\
& (0.8, 5) & 0.890 & 0.912 & 0.900 & 0.931 & $-2.87 \pm 1.21$ & $89.31 \pm 16.91$ \\
& (0.8, 10) & 0.941 & 0.955 & 0.948 & 0.966 & $-3.08 \pm 1.04$ & $90.85 \pm 15.87$ \\
& (0.8, 15) & 0.963 & 0.968 & 0.965 & 0.981 & $-2.98 \pm 1.14$ & $91.40 \pm 15.49$ \\
& (0.8, 20) & 0.964 & 0.968 & 0.966 & 0.985 & $-2.86 \pm 1.11$ & $91.72 \pm 15.26$ \\
& (0.8, 50) & 0.988 & 0.977 & 0.982 & 0.994 & \cellcolor{stratbest}$-2.86 \pm 0.93$$\star$ & $92.04 \pm 15.03$ \\
& (1.0, 3) & 0.973 & 0.976 & 0.975 & 0.990 & $-3.41 \pm 1.30$ & $90.20 \pm 16.52$ \\
& (1.0, 5) & 0.975 & 0.975 & 0.975 & 0.990 & $-3.43 \pm 1.56$ & $91.01 \pm 16.00$ \\
& (1.0, 10) & 0.991 & 0.989 & 0.990 & 0.998 & $-3.69 \pm 1.47$ & $91.70 \pm 15.49$ \\
& (1.0, 15) & 0.997 & 0.996 & 0.996 & 1.000 & $-3.46 \pm 1.44$ & $91.90 \pm 15.23$ \\
& (1.0, 20) & 0.996 & 0.989 & 0.993 & 0.997 & $-3.37 \pm 1.40$ & $91.92 \pm 15.23$ \\
& (1.0, 50) & 0.993 & 0.990 & 0.991 & 0.996 & $-3.07 \pm 1.14$ & \cellcolor{stratbest}$92.12 \pm 15.00$ \\
\midrule
\multirow{7}{*}{\rotatebox[origin=c]{90}{\scriptsize Top-$k$}}
& 1 & 0.961 & 0.981 & 0.971 & 0.988 & $-2.16 \pm 0.77$ & $57.82 \pm 41.59$ \\
& 3 & 0.930 & 0.940 & 0.934 & 0.967 & $-2.27 \pm 0.63$ & $75.78 \pm 27.15$ \\
& 5 & 0.919 & 0.932 & 0.925 & 0.957 & $-2.10 \pm 0.82$ & $82.40 \pm 20.66$ \\
& 10 & 0.885 & 0.890 & 0.887 & 0.929 & $-2.26 \pm 0.67$ & $87.06 \pm 15.68$ \\
& 15 & \cellcolor{overallbest}0.876 & \cellcolor{overallbest}\textbf{0.880}$\star$ & \cellcolor{overallbest}0.877 & \cellcolor{overallbest}0.924 & \cellcolor{stratbest}$-2.29 \pm 0.62$ & $87.18 \pm 16.37$ \\
& 20 & 0.898 & 0.886 & 0.891 & 0.929 & $-2.21 \pm 0.71$ & $87.82 \pm 15.78$ \\
& 50 & 0.924 & 0.925 & 0.924 & 0.955 & $-2.13 \pm 0.74$ & \cellcolor{stratbest}$88.29 \pm 16.12$ \\
\midrule
\multirow{5}{*}{\rotatebox[origin=c]{90}{\scriptsize Top-$p$}}
& 0.6 & 0.940 & 0.953 & 0.946 & 0.978 & $-2.21 \pm 0.66$ & $75.29 \pm 28.39$ \\
& 0.7 & 0.916 & 0.934 & 0.925 & 0.960 & $-2.26 \pm 0.62$ & $81.22 \pm 22.58$ \\
& 0.8 & 0.923 & 0.928 & 0.926 & 0.961 & $-2.22 \pm 0.67$ & $86.22 \pm 16.64$ \\
& 0.9 & 0.914 & 0.910 & 0.912 & 0.949 & $-2.23 \pm 0.66$ & $89.97 \pm 11.53$ \\
& 0.95 & \cellcolor{overallbest}0.891 & \cellcolor{overallbest}0.890 & \cellcolor{overallbest}0.890 & \cellcolor{overallbest}0.935 & \cellcolor{stratbest}$-2.28 \pm 0.60$ & \cellcolor{stratbest}$91.39 \pm 9.43$ \\
\midrule
\textbf{Human} & -- & -- & -- & -- & -- & $-2.73 \pm 0.11$ & $94.11 \pm 1.60$ \\
\bottomrule
\end{tabular}}
\caption{Detection performance and text properties for 48 selected non-temperature configurations, with $k\geq3$ for contrastive search. Predictability and diversity summarize the combined human and machine samples within each cell, not machine generations alone; they therefore should not be interpreted as generation-only statistics. Values highlighted in \colorbox{overallbest}{blue} indicate lowest detection rates within each strategy. Values highlighted in \colorbox{stratbest}{green} indicate configurations closest to human baseline values. Best overall configurations marked with $\star$. Human baseline shown for comparison.}
\label{tab:performance}
\end{table*}

\subsection{Effect of Architecture and Scale on AUC-ROC}
\label{sec:effect_architecture_scale}
\begin{figure*}[tp]
\centering
  \includegraphics[width=0.8\textwidth]{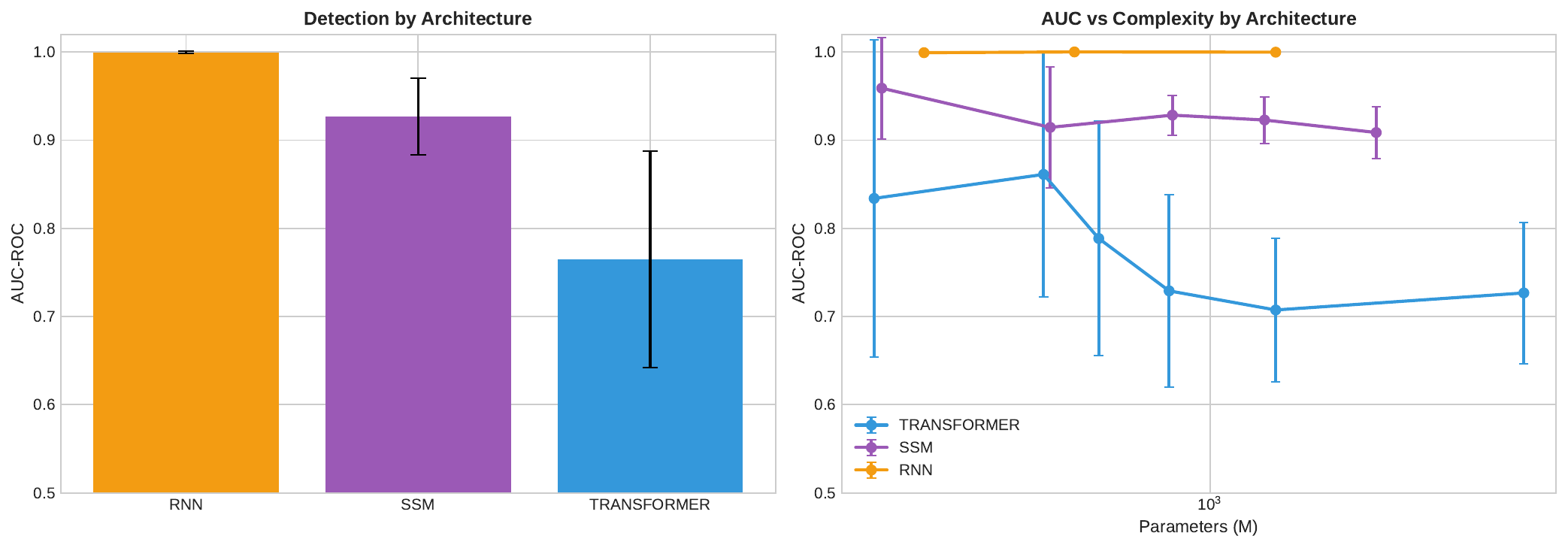}
  \caption{Effect of architecture and scale on AUC-ROC, comparing Transformer- and Non-Transformer-based architectures at different sizes.}
  \label{fig:architecture_aucroc}
\end{figure*}

To isolate the effects of model scale and architecture on detectability, we conducted a hierarchical regression analysis controlling for confounding factors; Figure~\ref{fig:architecture_aucroc} visualizes the resulting effects.

\paragraph{Scale Analysis.} We regressed AUC-ROC on log-transformed parameter count while conditioning on model family, dataset, and decoding strategy. Table~\ref{tab:scale_regression} reports the results. The coefficient of $\log(\text{params})$ is not statistically significant in family-specific regressions for three of four model families: Qwen2 ($\beta = -0.020$, $p = 0.229$), Mamba ($\beta = -0.013$, $p = 0.108$), and RWKV ($\beta = 0.000$, $p = 0.468$). Only GPT2 shows a significant within-family effect ($\beta = -0.052$, $p = 0.015$). Similar patterns hold for Logistic Regression and Naive Bayes classifiers. These results indicate that \emph{within model families}, increasing scale does not reliably reduce detectability.

\paragraph{Architecture Analysis.} Controlling for scale, dataset, and decoding strategy, we find significant differences across architectures. Non-Transformer architectures (Mamba, RWKV) exhibit significantly higher detectability than Transformer baselines ($\Delta\text{AUC} = 0.180$, $p < 0.001$). This suggests that high detectability is \emph{not unique to Transformer architectures}; rather, it may persist in non-Transformer models when generation relies on likelihood-based truncation.

\begin{table}[H]
\centering
\small
\resizebox{\columnwidth}{!}{%
\begin{tabular}{lccc}
\toprule
\textbf{Predictor} & \textbf{$\beta$} & \textbf{SE} & \textbf{$p$-value} \\
\midrule
\multicolumn{4}{l}{\textit{Scale Effect (within-family)}} \\
$\log(\text{params})$ & $-0.021$ & $0.009$ & $0.017^{\dagger}$ \\
\midrule
\multicolumn{4}{l}{\textit{Architecture Effect (controlling for scale)}} \\
Non-Transformer vs.\ Transformer & $+0.180$ & $0.018$ & $<0.001$ \\
\bottomrule
\end{tabular}}
\caption{Regression coefficients for scale and architecture effects on AUC-ROC (Random Forest classifier). $^{\dagger}$Pooled estimate; family-specific regressions show non-significant effects for 3 of 4 families (Qwen2: $p=0.229$; Mamba: $p=0.108$; RWKV: $p=0.468$).}
\label{tab:scale_regression}
\end{table}

\paragraph{Interpretation.} Detectability persists across the evaluated model families, but both the pooled scale coefficient and the architecture coefficient are statistically significant. Decoding and model characteristics both contribute to the observed variation in detectability.

\subsection{Robustness to the Choice of Reference Model}
\label{sec:robustness}

Our analysis relies on predictability estimates computed using a reference language model (OPT-2.7B). To assess whether our findings depend on this specific choice, we evaluate agreement across multiple judge models spanning different architectures and scales, including OPT-2.7B, Llama-3.1-8B, Qwen2-7B, and GPT2-medium. For each model, we compute predictability scores over the same set of human and machine-generated texts and measure pairwise Spearman rank correlations. As shown in Table~\ref{tab:judge_correlation}, we observe consistently high correlations across all model pairs ($\rho \geq 0.88$), indicating strong agreement in the relative ordering of texts by predictability.

\begin{table}[H]
\centering
\scriptsize
\setlength{\tabcolsep}{6pt}
\renewcommand{\arraystretch}{1.2}
\resizebox{\columnwidth}{!}{%
\begin{tabular}{lcccc}
\toprule
& \textbf{OPT-2.7B} & \textbf{Llama-3.1-8B} & \textbf{Qwen2-7B} & \textbf{GPT2-medium} \\
\midrule
\textbf{OPT-2.7B}     & 1.00 & 0.93 & 0.91 & 0.88 \\
\textbf{Llama-3.1-8B}   & 0.93 & 1.00 & 0.94 & 0.90 \\
\textbf{Qwen2-7B}      & 0.91 & 0.94 & 1.00 & 0.89 \\
\textbf{GPT2-medium}  & 0.88 & 0.90 & 0.89 & 1.00 \\
\bottomrule
\end{tabular}}
\caption{Spearman rank correlation ($\rho$) between predictability scores computed using different judge models.}
\label{tab:judge_correlation}
\end{table}

\subsection{Case Study}
\label{sec:case_study}

Table~\ref{tab:case_study_3} extends the case study of Section~\ref{sec:quality_dissociation} with longer excerpts on the same prompt.

\begin{table*}[tp]
\begin{center}
\resizebox{0.95\textwidth}{!}{
    \begin{tabular}{|p{3.5cm}|p{16.3cm}|p{2.1cm}|}
\hline
\textbf{Decoding Method} & \textbf{Prompt: "Headlam served as Officer Commanding North @-@ Western Area in 1946, and as Director of Training from 1947 to 1950. In 1950 - 51,"} & \textbf{P(Machine)}\\ \hline\hline

Qwen 2 CS (1.0, 20) & \colorbox{erratic}{" Headelm oversa  Operation Agulhene535af69, to reoccupyo portug 1 st colonial outst a ncesin}
\colorbox{erratic}{AFRicacoun trie sto preva l sepa rabli st mov me s . In th es  on he wro 4he couthoea2 rsh ipp ublicatin}
\colorbox{erratic}{hitherto re t aine o secret}... (text continues)" & 0.88\\ \hline

Llama-3.1 beam (3) & "he served as Commanding Officer of the 1st Battalion, The Queen's Royal Regiment (West Yorkshire Regiment). In 1951 2013 52, he served as Commanding Officer of the 2nd Battalion, \colorbox{repetitive}{The Queen's Royal Regiment (West Yorkshire Regiment). In 1952 2013 53, he served as Commanding}
 \colorbox{repetitive}{Officer of the 3rd Battalion, The Queen's Royal Regiment (West Yorkshire Regiment). In 1953 2013 54} \colorbox{repetitive}{, he served as Commanding Officer of the 4th Battalion,} ... (text continues)" & 1.00\\ \hline \hline

\end{tabular}
}
\caption{Case study comparing text quality and detectability. The probability of machine authorship ($P(\text{Machine})$) is estimated by Random Forest. Degenerate (repetitive) text is highlighted in \colorbox{repetitive}{purple}; erratic (incoherent) text is highlighted in \colorbox{erratic}{brown}. The examples illustrate different forms of poor-quality machine output.}
\label{tab:case_study_3}
\end{center}
\end{table*}

\subsection{Contrastive Search Parameter Interactions}
\label{sec:cs_interactions}

\begin{figure*}[tp]
\centering
  \includegraphics[width=0.88\textwidth,height=0.26\textheight,keepaspectratio]{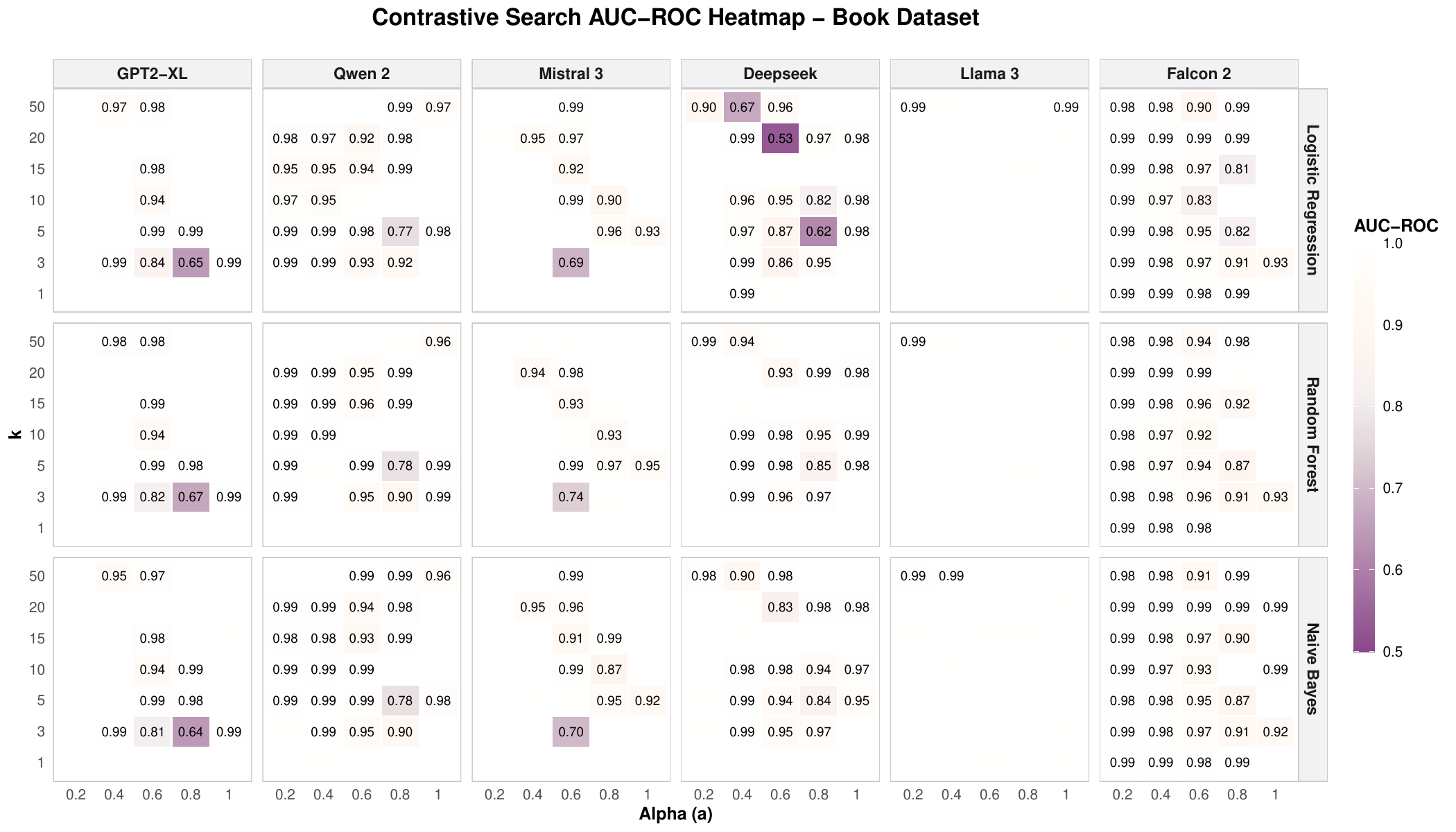}\\[2pt]
  \includegraphics[width=0.88\textwidth,height=0.26\textheight,keepaspectratio]{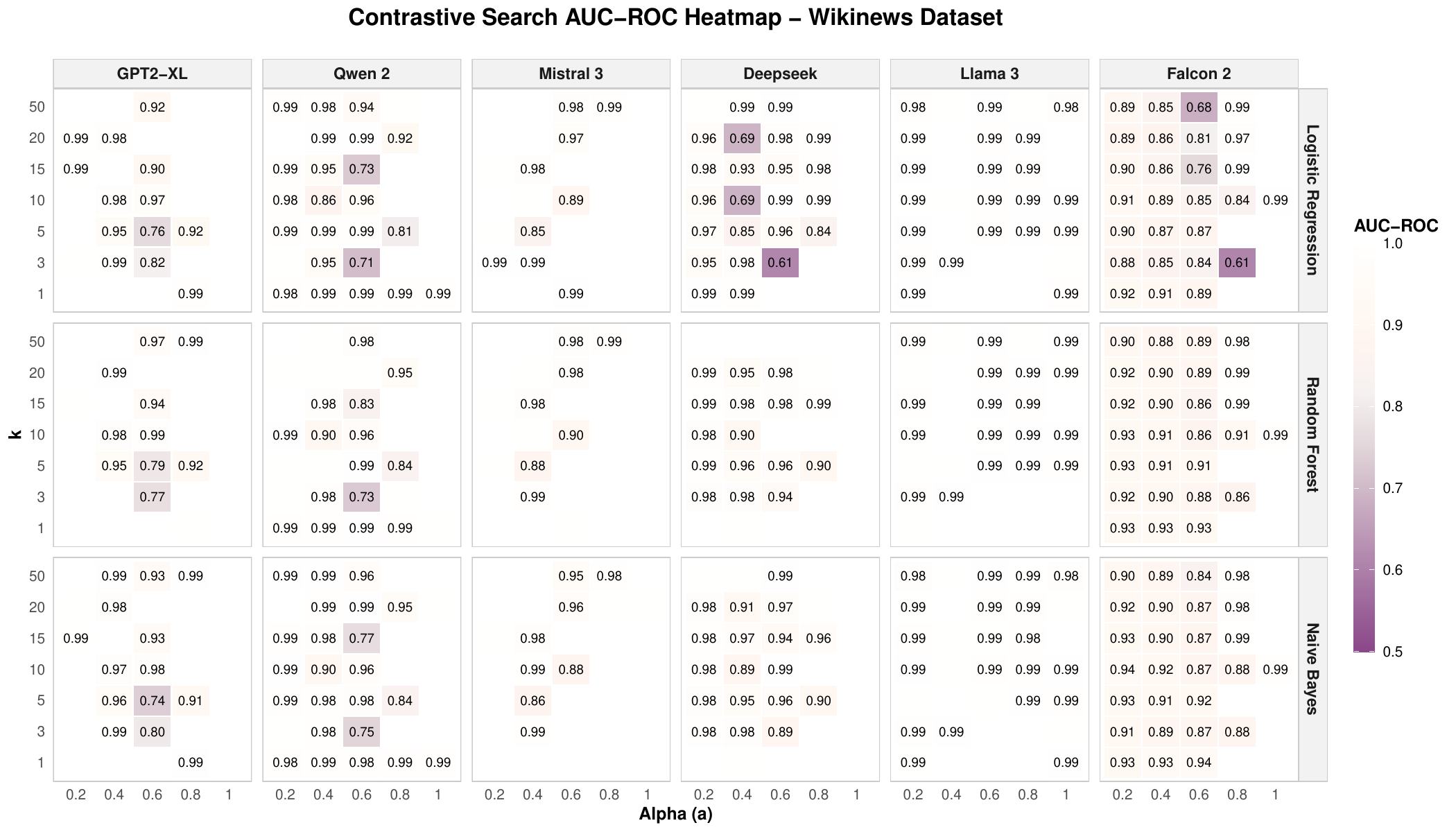}\\[2pt]
  \includegraphics[width=0.88\textwidth,height=0.26\textheight,keepaspectratio]{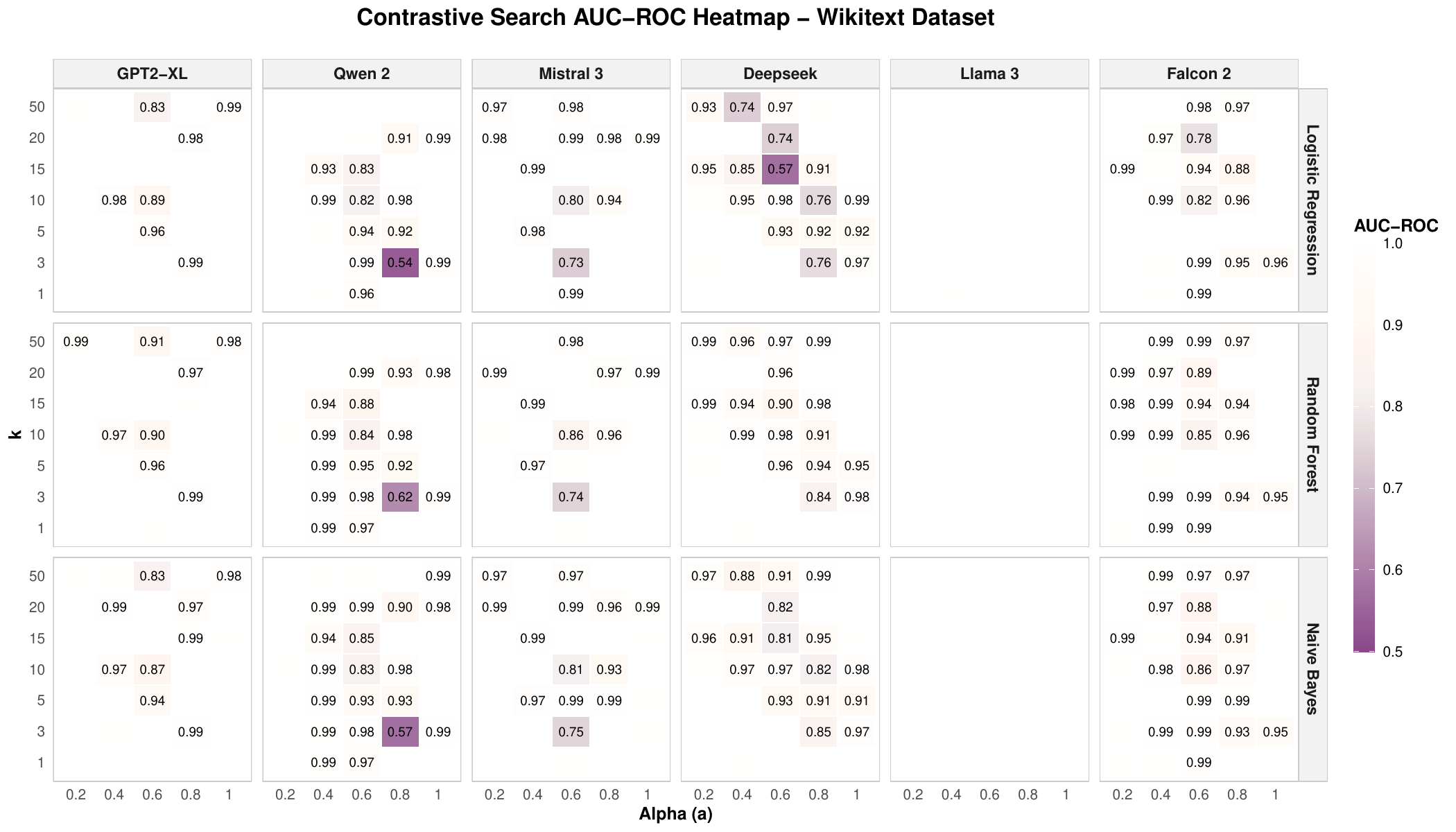}
  \caption{AUC-ROC heatmaps for contrastive search showing interactions between $k$ and $\alpha$ parameters. Darker regions indicate lower detectability. Facet labels ``Llama 3'' and ``Mistral 3'' denote the Llama-3.1 and Mistral-7B checkpoints of Section~\ref{sec:experimental_setup}.}
  \label{fig:cs_heatmaps}
\end{figure*}

Figure~\ref{fig:cs_heatmaps} shows AUC-ROC heatmaps for contrastive search, illustrating the interaction between the $k$ and $\alpha$ parameters across all three datasets. Darker regions indicate lower detectability; the least detectable cells concentrate at intermediate $\alpha$ with moderate $k$, consistent with Table~\ref{tab:least_distinguishable}.

\subsection{AUC-ROC Evolution Across Configurations}
\label{sec:auc_evolution}

\begin{figure*}[tp]
\centering
  \includegraphics[width=0.82\textwidth,height=0.26\textheight,keepaspectratio]{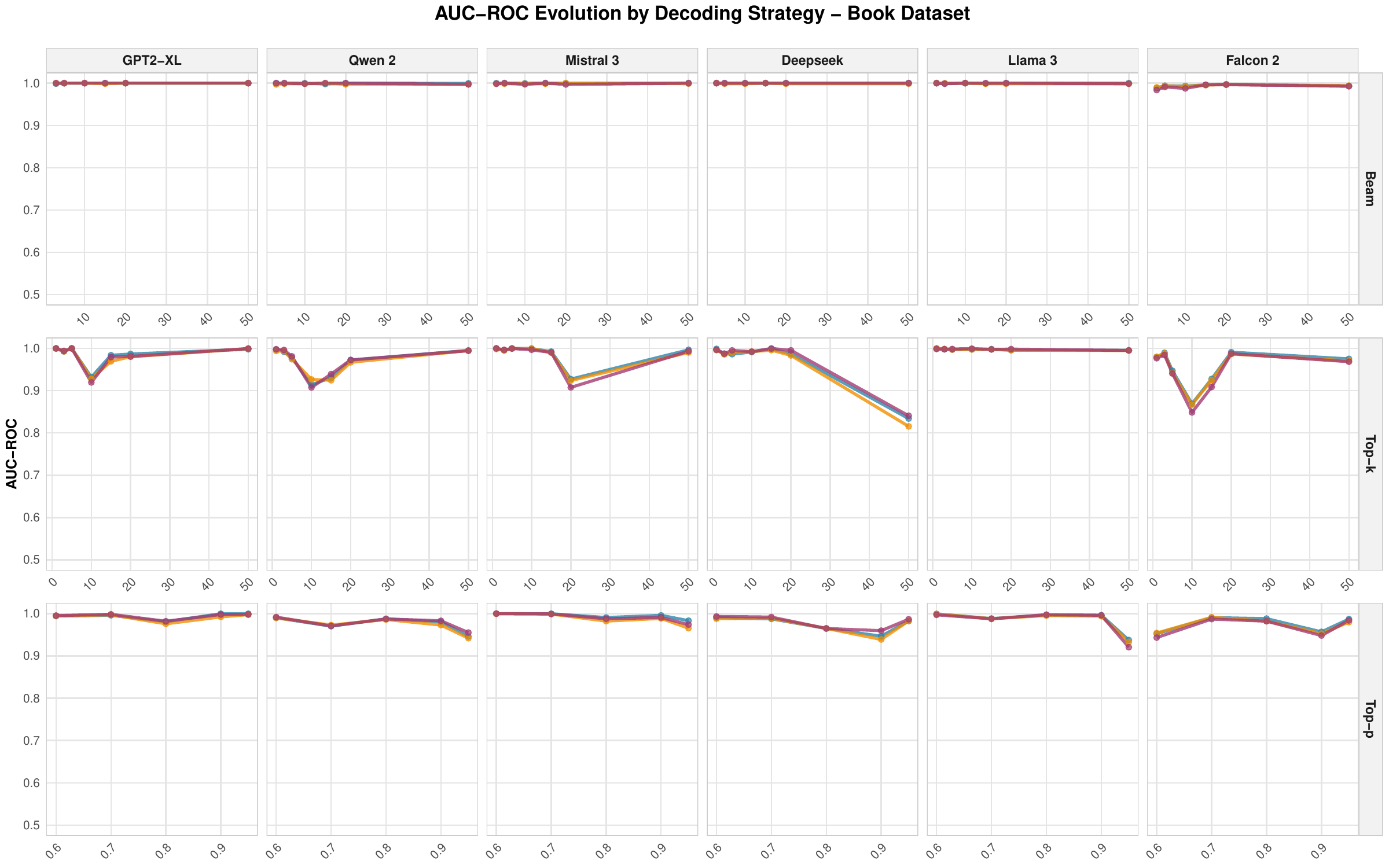}\\[2pt]
  \includegraphics[width=0.82\textwidth,height=0.26\textheight,keepaspectratio]{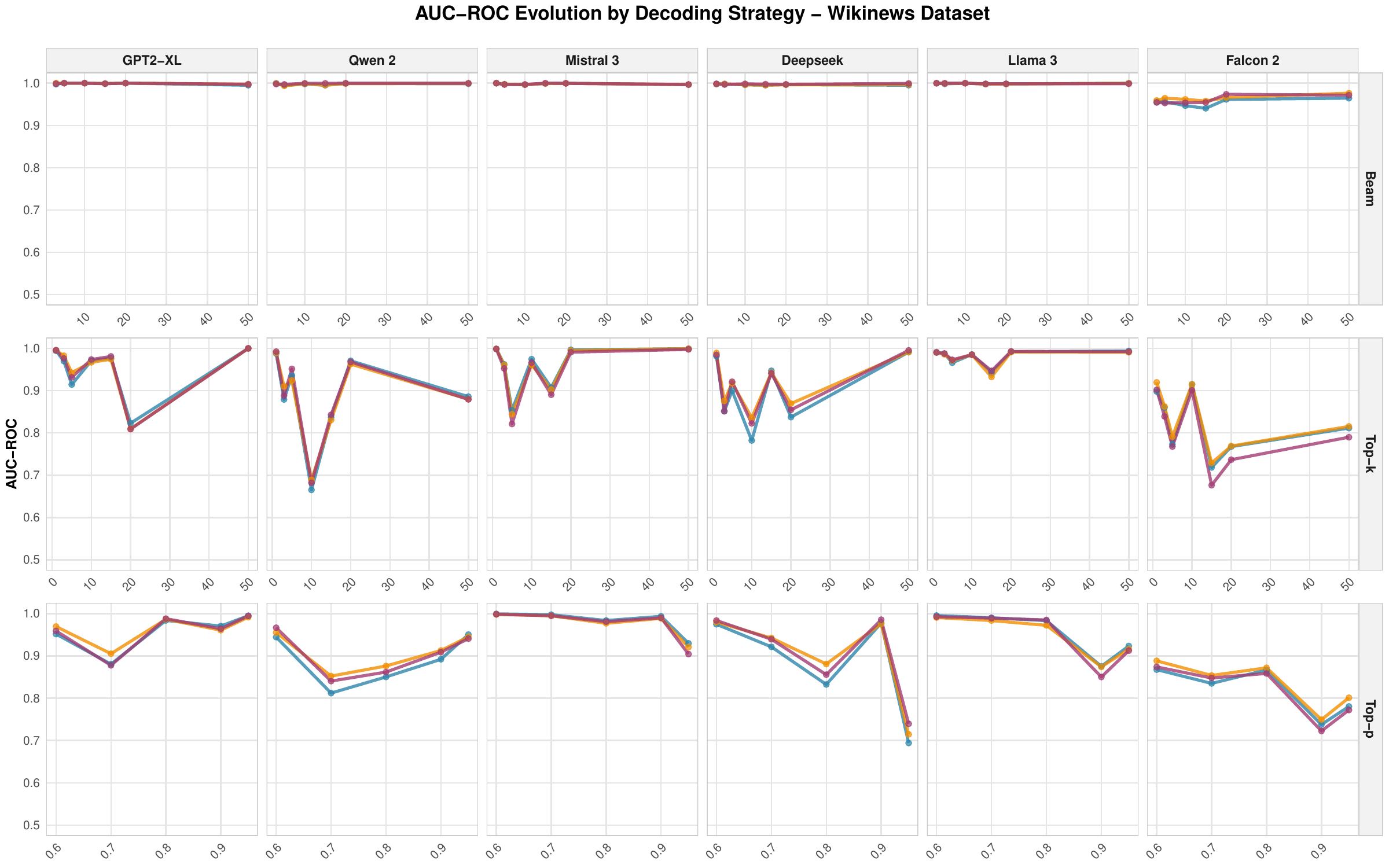}\\[2pt]
  \includegraphics[width=0.82\textwidth,height=0.26\textheight,keepaspectratio]{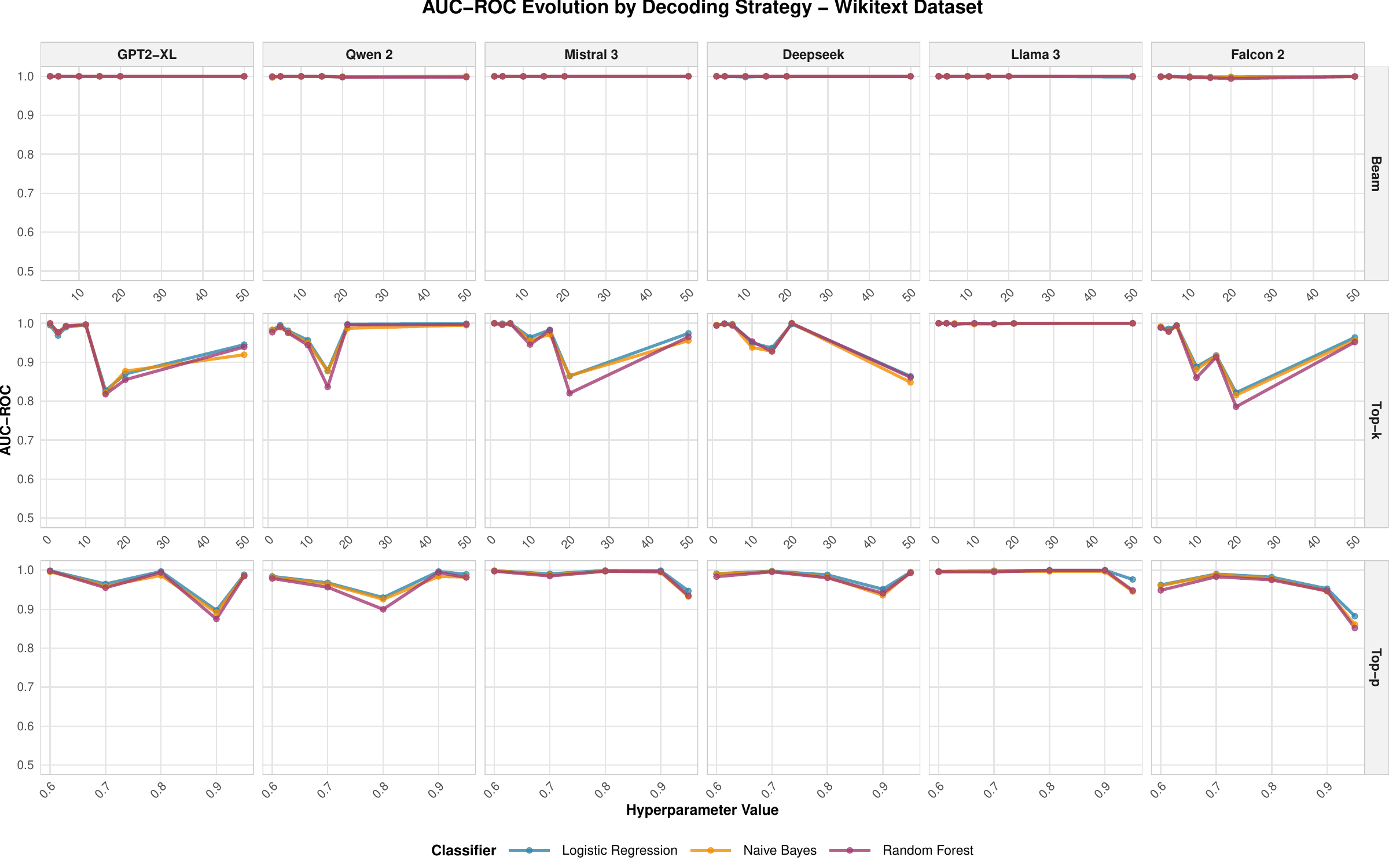}
  \caption{Evolution of AUC-ROC across hyperparameter values for multiple decoding strategies, models, and datasets. Larger $k$/$p$ values generally reduce detectability, consistent with the hypothesis that relaxing truncation reduces the distributional mismatch. Facet labels ``Llama 3'' and ``Mistral 3'' denote the Llama-3.1 and Mistral-7B checkpoints of Section~\ref{sec:experimental_setup}.}
  \label{fig:auc_evolution}
\end{figure*}

Figure~\ref{fig:auc_evolution} traces the evolution of AUC-ROC across hyperparameter values for multiple decoding strategies; the variation across settings shows how decoding choices affect detectability.

\subsection{Token Set Overlap at Different Truncation Levels Across Varying Scale}
\label{sec:overlap}

\begin{figure*}[tp]
\centering
  \includegraphics[width=0.66\textwidth,height=0.32\textheight,keepaspectratio]{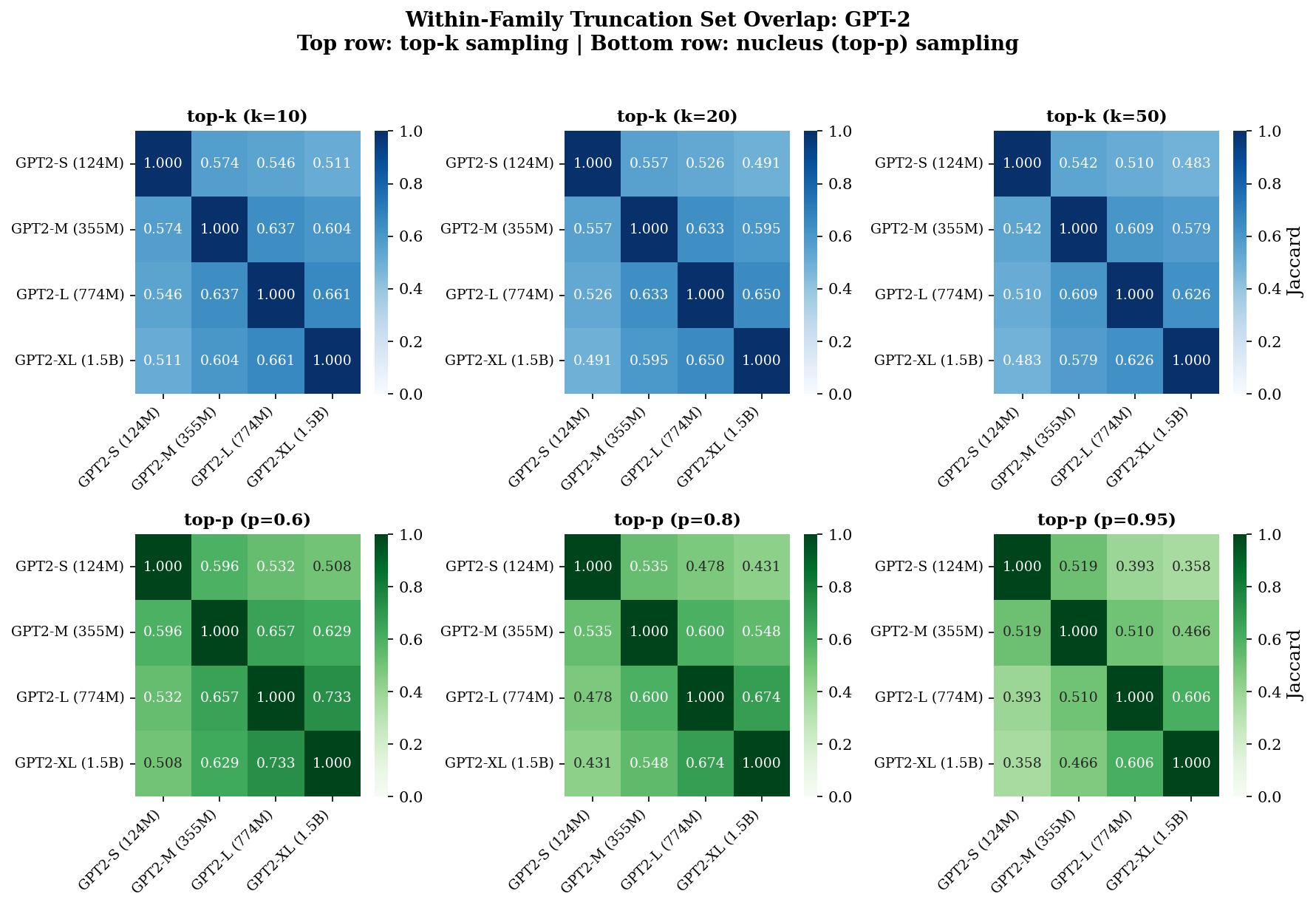}\\[2pt]
  \includegraphics[width=0.66\textwidth,height=0.32\textheight,keepaspectratio]{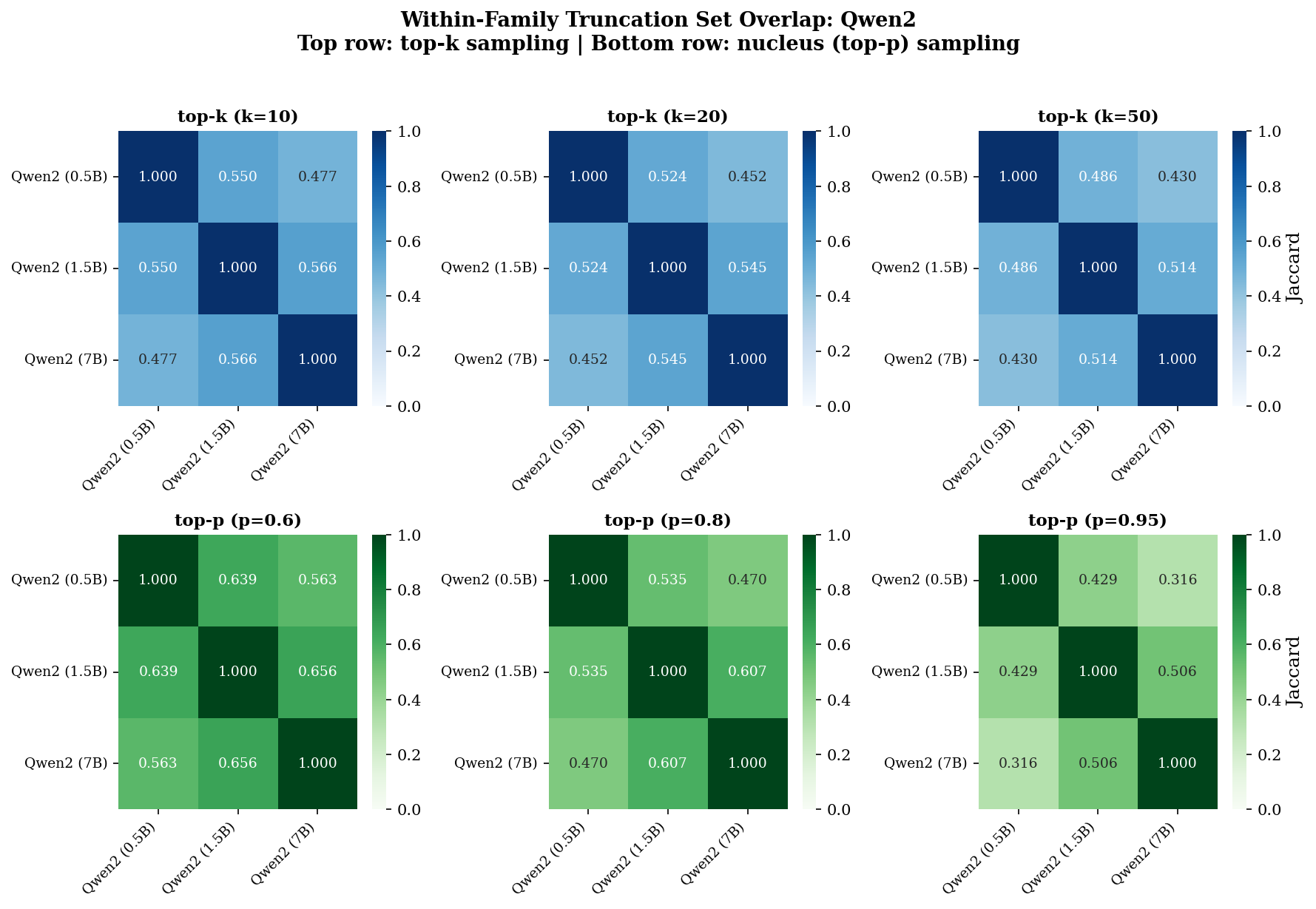}
  \caption{Within-family truncation set overlap, as measured by the Jaccard distance. \textbf{Top:} Example for GPT2 (from 124M to 1.5B). \textbf{Bottom:} Example for Qwen2 (from 0.5B to 7B) at different truncation schemes: \textbf{Blue:} Top-$k$ sampling with $k \in \{10, 20, 50\}$. \textbf{Green:} Top-$p$ sampling with $p \in \{0.6, 0.8, 0.95\}$.}
  \label{fig:overlap}
\end{figure*}

Figure~\ref{fig:overlap} visualizes within-family truncation-set overlap using the Jaccard distance, referenced from Sections~\ref{sec:model_results} and~\ref{sec:cross_model}.

\paragraph{Overlap Discussion.} We examine whether models within the same architectural family converge on similar truncation sets. Overlap describes how candidate sets change across model sizes; it does not by itself measure recovery of human-selected tokens. Within-family overlap is consistently high across both families. At $k=10$, GPT-2 models exhibit a mean overlap of 0.589 ($\pm$0.057), while Qwen2 models show 0.531 ($\pm$0.058). Overlap decreases with the size gap between models: for GPT-2, adjacent sizes achieve the highest overlap (0.661 for GPT2-L to GPT2-XL), while the largest gap yields the lowest (0.511 for GPT2-S to GPT2-XL). The same pattern holds for Qwen2, with overlap ranging from 0.566 (adjacent) to 0.477 (largest gap). This gradient indicates that scaling induces gradual distributional drift, but does not fundamentally restructure which tokens receive high probability. Under nucleus sampling, overlap decreases as $p$ increases, reflecting the expanding nucleus size: at $p=0.95$, mean overlap is 0.475 (GPT-2) and 0.417 (Qwen2), yet remains substantial.

\subsection{Detectability--Quality Dissociation}
\label{sec:appendix_quality}

\begin{figure*}[tp]
\centering
  \includegraphics[width=0.7\textwidth]{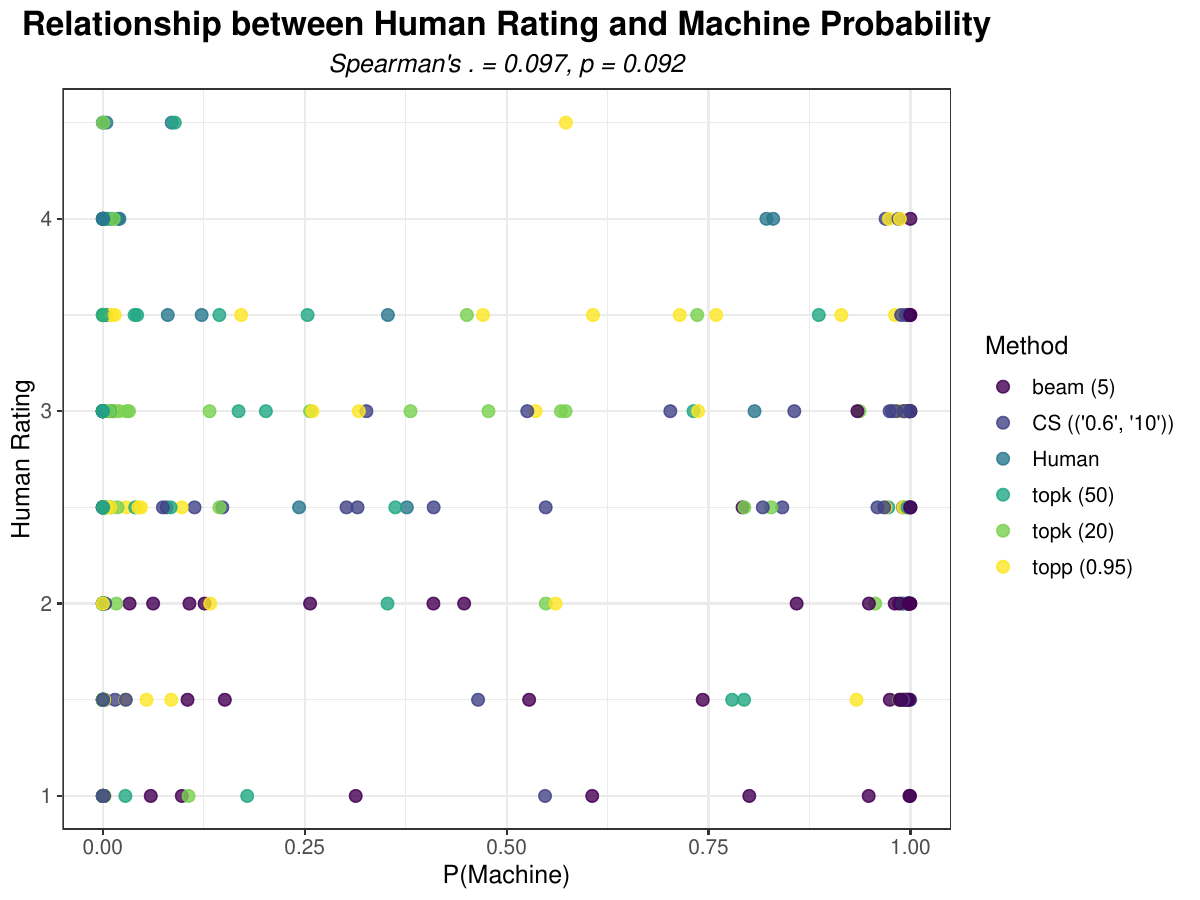}
  \caption{Relationship between human quality ratings (y-axis; Likert scale 1--5) and the estimated probability of machine-generated text (x-axis) across different decoding strategies. The association is weak, with a very small and statistically non-significant Spearman correlation. Human-written texts receive the highest ratings, followed by Top-$p$ sampling, and Top-$k$ sampling. Beam search yields the lowest ratings, with contrastive search performing slightly better. Results were computed from publicly available human evaluation results by \citet{garces-arias-etal-2025-statistical}.}
  \label{fig:quality}
\end{figure*}

Figure~\ref{fig:quality} plots human quality ratings against estimated machine probability, quantifying the dissociation reported in Section~\ref{sec:quality_dissociation}.

\subsection{Linguistic Composition of the Blind Spot}
\label{sec:pos_analysis}

\paragraph{Sample and checkpoints.} We sampled 200 unique prompt/reference pairs without replacement from each of the BookCorpus, WikiNews, and WikiText reference checkpoints in our public repository. The available files contain 1,939, 2,000, and 1,300 unique pairs, respectively; this POS subset is distinct from the full detection corpus. The saved input manifest identifies every selected pair. We use \texttt{gpt2}, \texttt{gpt2-medium}, and \texttt{gpt2-large} (124M, 355M, and 774M parameters), with float32 inference. Their shared tokenizer holds subword segmentation fixed across the three sizes, allowing a within-family analysis. Extending the analysis to other model families remains future work.

\paragraph{Scoring and alignment.} We remove an explicit leading \texttt{<|endoftext|>} marker from the prompt, concatenate the original prompt and reference, and tag the combined text with spaCy 3.8.7, \texttt{en\_core\_web\_sm} 3.8.0. A fast GPT-2 tokenizer supplies character offsets. We score at most 256 continuation subwords under the full preceding prompt and continuation context; the first unconditioned token, if any, is not scored. A subword receives a POS tag only when all its non-whitespace characters belong to one spaCy token. Whitespace-only tokens and subwords spanning multiple words are excluded from POS denominators. Ranks use the conservative tie convention of Section~\ref{sec:preliminaries}.

\paragraph{Alignment audit.} Across the three models, 353,808 continuation-token observations were selected and 346,518 were aligned: 4,689 whitespace-only observations and 2,601 multiword observations were omitted. There were no unmapped or prompt-boundary-crossing observations. Multiword omissions represent 0.55--1.01\% of selected tokens by domain; the GPT-2 variants share these alignments. No ties straddled the $k=10$ or $k=50$ boundaries. As a conservative sensitivity check, assigning every omitted multiword observation to the function group and treating it as excluded reduces the pooled $k=10$ ratio to 3.07, leaving its direction unchanged.

\begin{figure*}[tp]
\centering
  \includegraphics[width=0.65\textwidth]{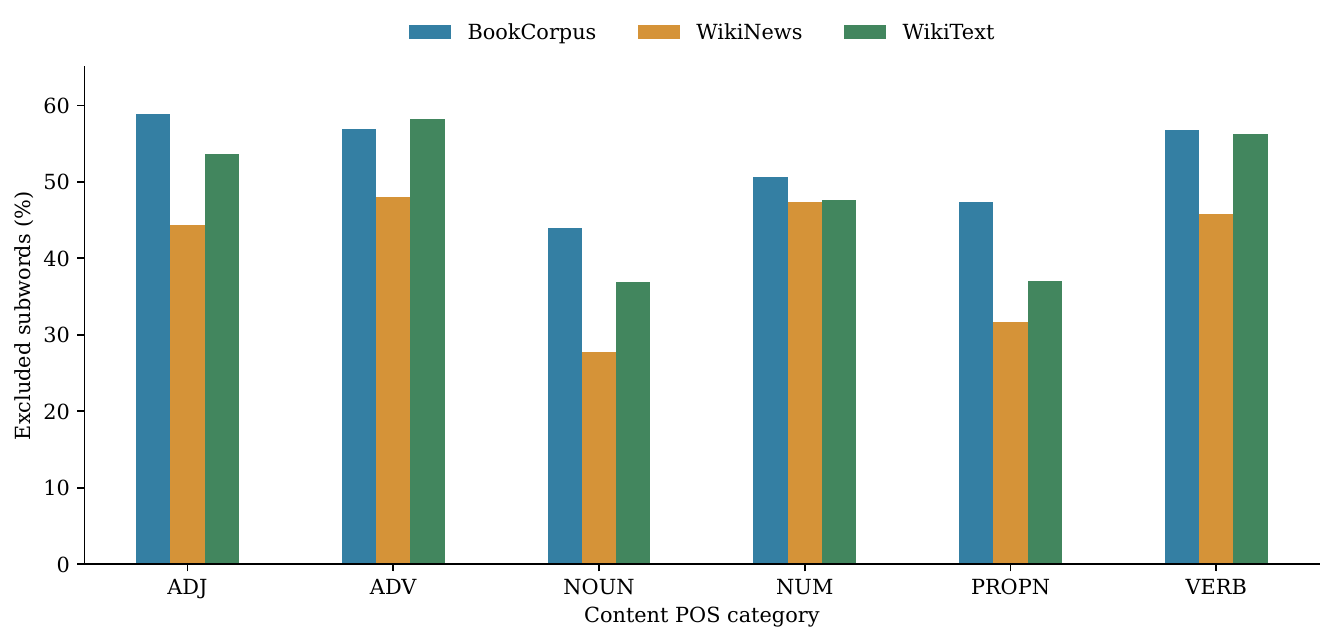}
  \caption{Content-category subword exclusion at $k=10$ by benchmark domain, pooling counts across the three GPT-2 models. Each domain contributes 200 prompt/reference pairs; the number of aligned subwords varies by domain.}
  \label{fig:pos_domain}
\end{figure*}

\paragraph{Groups and aggregation.} Content categories are NOUN, VERB, ADJ, ADV, PROPN, and NUM; function categories are DET, ADP, CCONJ, SCONJ, PRON, AUX, and PART. Punctuation, interjections, symbols, and other tags form a separate group. Reported and plotted rates pool excluded and total aligned subword counts across the fixed model set and sampled pairs. Thus, a word split into multiple subwords contributes multiple observations. A supplementary macro-average, weighting observed POS/model/domain rates equally, gives 47.2\% versus 12.8\% and a ratio of 3.69 at $k=10$; this differs from the pooled ratio of 3.53.

\paragraph{Uncertainty and consistency.} The 95\% percentile interval uses 1,000 bootstrap resamples of prompt/reference pairs within each domain (seed 42); each resampled pair carries its observations for all three models and both word groups. It quantifies passage-sampling uncertainty conditional on these models and automatic tags, not uncertainty over model families or original source documents. Pooled exclusion is higher for content than function subwords in every model/domain cell (Table~\ref{tab:pos_model_domain}).

\paragraph{Category and domain patterns.} At $k=10$, adverbs (54.5\%), verbs (52.5\%), adjectives (52.2\%), and numerals (47.7\%) have the highest exclusion rates among the designated content categories. Proper nouns and nouns have rates of 35.4\% and 35.2\%. Function-category rates range from 6.7\% for particles to 26.0\% for subordinating conjunctions. Interjections have a higher rate but only 111 model-token observations and are outside the designated content/function groups.\\Figure~\ref{fig:pos_domain} shows domain variation: BookCorpus has the highest pooled exclusion for adjectives, nouns, proper nouns, and verbs, while WikiText is highest for adverbs.

\paragraph{Category prevalence.} Across 17 observed POS categories, category prevalence (the share of aligned subword observations) has little linear association with pooled exclusion ($r=-0.094$, $p=0.719$; Figure~\ref{fig:pos_analysis}C). A common category can contain many rare words, so this aggregate comparison does not rule out lexical-frequency effects. 

\begin{table}[H]
\centering
\scriptsize
\setlength{\tabcolsep}{3pt}
\begin{tabular}{llrrr}
\toprule
Model & Domain & Content & Function & Ratio \\
\midrule
GPT-2 small & Book & 55.5 & 14.7 & 3.77 \\
 & News & 40.5 & 11.7 & 3.46 \\
 & Wiki & 48.4 & 13.0 & 3.73 \\
GPT-2 medium & Book & 50.3 & 13.3 & 3.78 \\
 & News & 34.9 & 10.7 & 3.27 \\
 & Wiki & 42.4 & 11.6 & 3.64 \\
GPT-2 large & Book & 48.9 & 12.7 & 3.84 \\
 & News & 32.5 & 9.8 & 3.33 \\
 & Wiki & 39.8 & 10.7 & 3.73 \\
\bottomrule
\end{tabular}
\caption{Pooled subword exclusion rates (\%) at $k=10$ within each GPT-2 model/domain cell. Ratios are computed before rounding.}
\label{tab:pos_model_domain}
\end{table}

\end{document}